\documentclass[10pt,twocolumn,letterpaper]{article}

\usepackage{cvpr}              % To produce the CAMERA-READY version
\usepackage{bm}
\usepackage{color}
\usepackage[table]{xcolor}  % 支持表格上色
\usepackage{multicol}  % 在导言区添加
\usepackage{pgfplots}
\pgfplotsset{compat=1.17}
\usepackage{xspace}
\usepackage{multirow}
\usepackage{graphicx} 
\usepackage{tabularx,booktabs}
\newcommand{\method}{$<$answer$>$\xspace}
\newcommand{\plusours}{ \ \ $\mathbf{+}$ \textbf{\method}}
\definecolor{ours}{RGB}{244,237,252}
\usepackage{bbding}
\usepackage{xcolor}
\definecolor{mygray}{RGB}{0,128,0}
\definecolor{upcolor}{RGB}{57,182,74}
\newcommand{\up}[1]{\textcolor{upcolor}{$\uparrow$ #1}}

\definecolor{cvprblue}{rgb}{0.21,0.49,0.74}
\usepackage[pagebackref,breaklinks,colorlinks,allcolors=cvprblue]{hyperref}

\def\paperID{19574} % *** Enter the Paper ID here
\def\confName{CVPR}
\def\confYear{2026}

\title{Can a Second-View Image Be a Language? Geometric and Semantic Cross-Modal Reasoning for X-ray Prohibited Item Detection}

\author{
Chuang Peng$^{1}$, Renshuai Tao$^{1}$\thanks{Corresponding author.} , Zhongwei Ren$^{1}$, Xianglong Liu$^{2}$, Yunchao Wei$^{1}$\\
\textsuperscript{1}{Institute of Information Science, Beijing Jiaotong University}\\
\textsuperscript{2}{State Key Laboratory of Complex \& Critical Software Environment, Beihang University}\\
\fontsize{8.5pt}{\baselineskip}\selectfont \tt mr.pengc@foxmail.com, rstao@bjtu.edu.cn\\}

\begin{document}
\maketitle

\begin{abstract}
% Automated X-ray threat detection is fundamentally hampered by the visual ambiguity and occlusion inherent in single-view systems, failing to leverage the geometric complementarity of standard dual-view setups. While recent Vision-Language Models (VLMs) introduce semantic reasoning, no unified framework exists that simultaneously bridges visual, geometric (cross-view), and linguistic modalities due to the scarcity of structured multimodal data. We propose a novel paradigm by formalizing the auxiliary X-ray view as a structured ``language-like modality" to enrich reasoning. To enable this, we construct DualXrayCap, the first large-scale dual-view caption corpus encoding hierarchical geometric and semantic correspondences, and derive a Supervised Fine-Tuning (SFT) dataset using an explicit Chain-of-Thought (CoT) structure: $\text{\textless top\textgreater , \textless side\textgreater , \textless conclusion\textgreater}$. Leveraging this resource, we introduce DualXrayBench, the first traceable benchmark for dual-view X-ray inspection, featuring eight diagnostic tasks to rigorously assess cross-view reasoning. Finally, our proposed model, GSR, achieves substantial improvements across all benchmark tasks, demonstrating superior robustness and generalization by jointly utilizing textual semantics and the auxiliary view's geometric constraint. This work establishes the foundation for geometrically and semantically grounded X-ray security systems.
Automatic X-ray prohibited items detection is vital for security inspection and has been widely studied. Traditional methods rely on visual modality, often struggling with complex threats. While recent studies incorporate language to guide single-view images, human inspectors typically use dual-view images in practice. This raises the question: \textbf{can the second view provide constraints similar to a language modality?} In this work, we introduce DualXrayBench, the first comprehensive benchmark for X-ray inspection that includes multiple views and modalities. It supports eight tasks designed to test cross-view reasoning. In DualXrayBench, we introduce a caption corpus consisting of 45,613 dual-view image pairs across 12 categories with corresponding captions. Building upon these data, we propose the \textbf{G}eometric (cross-view)-\textbf{S}emantic (cross-modality) \textbf{R}easoner (GSR), a multimodal model that jointly learns correspondences between cross-view geometry and cross-modal semantics, treating the second-view images as a ``language-like modality". To enable this, we construct the GSXray dataset, with structured Chain-of-Thought sequences: $\text{\textless top\textgreater , \textless side\textgreater , \textless conclusion\textgreater}$. Comprehensive evaluations on DualXrayBench demonstrate that GSR achieves significant improvements across all X-ray tasks, offering a new perspective for real-world X-ray inspection.%\footnote{The code is in the Supplementary Materials and will be open-sourced.}
% Using this corpus, we construct the Geometric (cross-view)-Semantic (cross-modal) (GSXray) dataset, designed to train Large Language Models (LLMs) with a Chain-of-Thought (CoT) structure:$\text{\textless top\textgreater , \textless side\textgreater , \textless conclusion\textgreater}$.

% Inspired by the notion that "images are also a language," we explore whether the second view can offer complementary geometric constraints akin to linguistic guidance.

% Therefore, in this paper, we present DualXrayCap, the first large-scale dual-view caption corpus that encodes hierarchical geometric and semantic correspondences. We also construct a Supervised Fine-Tuning (SFT) dataset utilizing a Chain-of-Thought (CoT) structure: $\text{\textless top\textgreater , \textless side\textgreater , \textless conclusion\textgreater}$. Based on this resource, we introduce DualXrayBench, the first traceable benchmark for dual-view X-ray inspection, featuring eight diagnostic tasks to rigorously evaluate cross-view reasoning. Our proposed model, GSR, achieves significant improvements across all benchmark tasks, demonstrating enhanced robustness and generalization by jointly leveraging textual semantics and the auxiliary view’s geometric constraints. This work lays the foundation for geometrically and semantically grounded X-ray security systems.
\end{abstract}

\begin{figure}[!t]
  \centering
\includegraphics[width=1\linewidth]{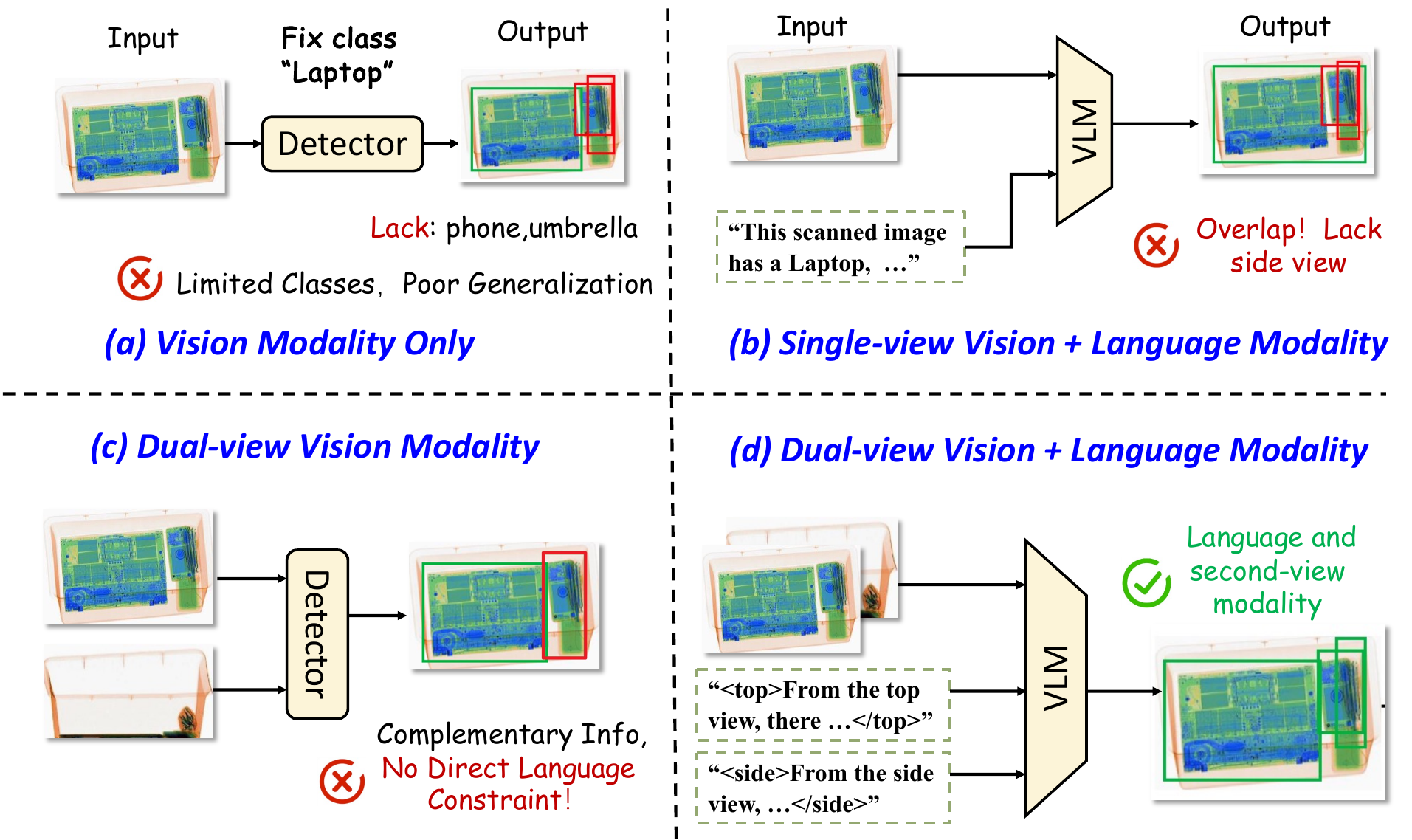}
\vspace{-0.2in}
   \caption{Comparison of existing X-ray prohibited-item detection strategies (a–c) with our DualXrayBench (d), which treats the second-view image as a language-like modality that provides additional constraints to enhance detection.}
   \label{fig:1}
   \vspace{-0.2in}
\end{figure}

% Comparison between existing X-ray prohibited item detection strategies (a, b, c) and our proposed DualXrayBench (d). The core idea is that the second-view image can be treated as a language-like modality in the X-ray detection task, providing additional constraints to enhance the detection performance.

\section{Introduction}
\label{sec:intro}

%安检很重要，需要人工智能技术已经迫切需要。
% With the acceleration of globalization and the continuous surge in international passenger and cargo traffic \cite{WSOL, Hixray}, ensuring safety and stability in high-traffic public transportation sectors, particularly aviation,  is a paramount global challenge, making X-ray baggage screening an indispensable security measure. However, the conventional process is fundamentally hampered by its heavy reliance on human operators, who are susceptible to visual fatigue and subjective errors when interpreting complex X-ray imagery containing overlapping and occluded objects \cite{akcay_survey,survey}. To effectively counter evolving threats and meet the pressing demand for enhanced screening efficiency, the integration of  artificial intelligence, specifically deep learning \cite{zeng2024collapsed,mallik2024priorband,guo2023cbanet,guo2023multidimensional}, for automated threat detection has become an urgent necessity in modern security operations.
%传统单模态的局限性
Automatic X-ray prohibited items detection \cite{GDXray,SIXray,OPIXray,Hixray,EDS} is crucial for security inspection and has been extensively studied. Traditional methods~\cite{PIXray, BalAffin_Jour, pidray_iccv} focus on extracting low-level visual features, such as edges and contours, to detect prohibited items. However, these systems struggle with real-world scenarios due to limited generalization and lack of semantic reasoning, making them unable to handle emerging threats or structural variations, such as distinguishing tablets from laptops~\cite{SIXray}. As a result, purely visual-modal-based models fail to provide robust detection.

The limitations of vision-only systems have sparked interest in integrating semantic reasoning through language modalities, driven by advancements in Vision-Language Models (VLMs) \cite{Clip_radford,liu2023llava,wasim2024vgdino}. This multi-modal approach enhances reasoning and generalization by grounding visual data in semantic space, allowing models to infer beyond pixel-level patterns. Recent studies have leveraged language to overcome the closed-set limitations of vision-only X-ray systems, using image-caption pairs and textual descriptors to regularize visual learning \cite{STING-BEE,PIXray-Caption}. While language-grounded models effectively address intra-class variance and improve open-vocabulary detection, dual-view X-ray imagery, a crucial modality used by human inspectors, has been overlooked. Human inspectors rely on dual views to recognize prohibited items despite occlusions and structural ambiguities. Inspired by the idea that ``images are also language", we had a sudden insight: \textbf{Can a second-view image be treated as a language, with its geometric correspondences jointly learned with linguistic semantics to enhance X-ray prohibited items detection?}

% Recent research has focused on fusing dual-view imagery to enhance detection under occlusion or overlap \cite{ldxray,dualray,dagnet}, but these methods remain vision-centric, neglecting semantic guidance. 

% While language-grounded models have proven effective in mitigating intra-class variance and improving open-vocabulary detection, a crucial modality routinely exploited by human inspectors remains largely overlooked. In practical security screening, operators rely on dual-view X-ray imagery to reconstruct a 3D model of baggage contents, thereby resolving occlusions and structural ambiguities that challenge single-view perception. In response, a complementary line of research has focused on exploiting dual-view X-ray imagery through specialized detection frameworks and curated datasets \cite{ldxray,dualray,dagnet}. These approaches aim to computationally fuse two orthogonal perspectives to enhance detection robustness, particularly under severe occlusion or complex object overlap. Crucially, existing dual-view methods remain largely vision-centric: they focus on fusing geometric and structural cues across views but do not exploit language-level semantic guidance. Motivated by this gap (\autoref{fig:1}), we ask: \textbf{Can a second view be regarded as a language—one whose geometric correspondences can be formalized as an additional modality and jointly learned with linguistic semantics to enhance robustness and accuracy in X-ray threat detection?}

%图像也是一种语言
Realizing this joint reasoning paradigm requires datasets that simultaneously capture semantic, geometric, and visual correspondences, but resources are largely absent in the X-ray domain. Existing single-view X-ray datasets enriched with image-caption pairs \cite{STING-BEE,PIXray-Caption} provide valuable semantic grounding but lack the geometric complementarity needed for cross-view reasoning. In contrast, emerging X-ray dual-view datasets \cite{ldxray,dualray} capture structural consistency between perspectives but remain vision-only, lacking linguistic alignment for semantic learning. As a result, no unified resource bridges visual, geometric, and linguistic modalities, limiting Vision-Language Models' ability to leverage the inherent dual-view consistency of X-ray imagery and generalizing in real-world security applications.

To address these challenges, in this work, we introduce \textbf{DualXrayBench}, the first comprehensive benchmark for X-ray prohibited items detection that integrates multiple views and modalities. DualXrayBench includes a dual-view caption corpus with 45,613 pairs of images across 12 categories, each accompanied by captions. Each sample provides: (1) image-level scene analyses for individual top and side views, (2) a holistic dual-view description combining complementary information, and (3) fine-grained object-level annotations with cross-view associations, including category, spatial relations, and occlusion relations. Moreover, DualXrayBench supports eight tasks designed to test cross-view reasoning, comprising 1,594 expert-verified visual question–answer pairs. Unlike conventional single-view evaluation settings, DualXrayBench employs a dual-view evaluation paradigm, assessing models' reasoning ability, cross-view consistency, and spatial understanding across eight diagnostic sub-tasks.

Building upon these data, we propose the \textbf{G}eometric-\textbf{S}emantic \textbf{R}easoner (\textbf{GSR}), a multimodal model that jointly learns textual semantics and dual-view geometric correspondences, treating the second-view image as a structured ``language-like modality". To enable this, we specifically construct the \textbf{GSXray dataset} with structured Chain-of-Thought (CoT) sequences: \textless top\textgreater, \textless side\textgreater, \textless conclusion\textgreater, explicitly modeling the reasoning flow from view-specific observations to unified semantic conclusions. Evaluations on DualXrayBench demonstrate that GSR achieves significant and consistent improvements across all tasks, showcasing superior cross-view reasoning, compositional understanding, and robustness to unseen prohibited items categories. The contributions are summarized as follows:
\begin{itemize}[noitemsep, left=0pt]

\item We are the first to propose that a second-view image can be treated as a language-like modality in X-ray detection task, providing extra constraints to enhance performance.
\item We introduce DualXrayBench, the first benchmark for X-ray detection with cross-view and cross-modal data, supporting eight tasks to assess cross-view reasoning, interpretability, consistency, and spatial understanding.
\item We propose the GSR method, a multimodal model that jointly learns correspondences between cross-view geometry and cross-modal semantics, achieving significant improvements across the eight X-ray detection tasks.

% \item \textbf{DualXrayCap:}We construct DualXrayCap, a large-scale dual-view caption corpus comprising 45613 paired images, which hierarchically encodes geometric and semantic correspondences across complementary views. Furthermore, we derive a structured SFT dataset featuring explicit Chain-of-Thought sequences \textless top\textgreater\ , \textless side\textgreater , \textless conclusion\textgreater to supervise the reasoning transition from view-specific observation to unified semantic inference.

% \item \textbf{DualXrayBench:}We establish DualXrayBench, a expert-verified benchmark designed for dual-view evaluation, offering 1.6K high-quality VQA pairs that systematically assess reasoning, spatial understanding, and cross-view consistency across eight diagnostic sub-tasks.

% \item \textbf{GSR:} Finally, We develop the Geometric–Semantic Reasoner(GSR), a multimodal reasoning model that unifies textual semantics and geometric correspondences between complementary views, achieving state-of-the-art performance and on DualXrayBench.
\end{itemize}
%strong generalization
% To systematically evaluate these capabilities, we further develop \textbf{DualXrayBench}, a multi-task diagnostic benchmark with 1,594 expert-verified VQA pairs. Unlike conventional datasets, DualXrayBench targets reasoning, interpretability, and cross-view consistency, probing models across eight diagnostic sub-tasks (e.g., Perceptual, Relational, Causal/Occlusion).

\begin{table}[!b]
\vspace{-0.1in}
\setlength{\tabcolsep}{0.36mm}
% \vspace{-0.7in}
  \fontsize{8.1}{13.8}\selectfont
  \centering
  \begin{tabular}{lccccccccc}
    \toprule
    Dataset & 2-view & Caption &Open & $N_{c}$ & $N_{p}$  & Task & Year  \\
    \midrule
    GDXray~\cite{GDXray} & \textcolor[rgb]{1,0,0}{\XSolidBrush} &  \textcolor[rgb]{1,0,0}{\XSolidBrush} &\textcolor[rgb]{0,0.5,0}{\Checkmark}& 3 & 8,150 & D & 2015 \\
    Compass XP~\cite{compassxp}& \textcolor[rgb]{1,0,0}{\XSolidBrush} & \textcolor[rgb]{1,0,0}{\XSolidBrush}&\textcolor[rgb]{0,0.5,0}{\Checkmark}&366&1928&C&2019  \\
    SIXray~\cite{SIXray}  & \textcolor[rgb]{1,0,0}{\XSolidBrush} & \textcolor[rgb]{1,0,0}{\XSolidBrush}& \textcolor[rgb]{0,0.5,0}{\Checkmark}&6 & 8,929 & C, D & 2019  \\
    OPIXray~\cite{OPIXray} &  \textcolor[rgb]{1,0,0}{\XSolidBrush} & \textcolor[rgb]{1,0,0}{\XSolidBrush} & \textcolor[rgb]{0,0.5,0}{\Checkmark}&5 & 8,885 & D  & 2020  \\
    PIDray~\cite{pidray_iccv} &  \textcolor[rgb]{1,0,0}{\XSolidBrush} &\textcolor[rgb]{1,0,0}{\XSolidBrush} & \textcolor[rgb]{0,0.5,0}{\Checkmark}&12 & 47,677 & D, S & 2021  \\
    HiXray~\cite{Hixray} & \textcolor[rgb]{1,0,0}{\XSolidBrush} & \textcolor[rgb]{1,0,0}{\XSolidBrush} & \textcolor[rgb]{0,0.5,0}{\Checkmark}&8 & 45,364  & D & 2021 \\
    PIXray~\cite{PIXray} & \textcolor[rgb]{1,0,0}{\XSolidBrush} & \textcolor[rgb]{1,0,0}{\XSolidBrush} & \textcolor[rgb]{0,0.5,0}{\Checkmark}&15 & 5,046 & D, S & 2022 \\ 
    CLCXray~\cite{CLCXray} & \textcolor[rgb]{1,0,0}{\XSolidBrush} & \textcolor[rgb]{1,0,0}{\XSolidBrush} & \textcolor[rgb]{0,0.5,0}{\Checkmark}&12 & 9,565  & D &  2022 \\
    EDS~\cite{EDS} & \textcolor[rgb]{1,0,0}{\XSolidBrush} & \textcolor[rgb]{1,0,0}{\XSolidBrush} & \textcolor[rgb]{0,0.5,0}{\Checkmark}&10 & 14,219  & D & 2022 \\
    FSOD~\cite{FSOD} & \textcolor[rgb]{1,0,0}{\XSolidBrush} & \textcolor[rgb]{1,0,0}{\XSolidBrush} & \textcolor[rgb]{0,0.5,0}{\Checkmark}&20 & 12,333  & D & 2022 \\
    LPIXray~\cite{LPIXray} & \textcolor[rgb]{1,0,0}{\XSolidBrush}  & \textcolor[rgb]{1,0,0}{\XSolidBrush} & \textcolor[rgb]{1,0,0}{\XSolidBrush} &18 & 60,950  & D & 2023 \\
    \midrule
    MV-Xray~\cite{mvxray} & \textcolor[rgb]{0,0.5,0}{\Checkmark} & \textcolor[rgb]{1,0,0}{\XSolidBrush} & \textcolor[rgb]{1,0,0}{\XSolidBrush} &2 & 3,231 & D & 2019 \\
    deei6~\cite{deei6} & \textcolor[rgb]{0,0.5,0}{\Checkmark} & \textcolor[rgb]{1,0,0}{\XSolidBrush} & \textcolor[rgb]{1,0,0}{\XSolidBrush} &6 & 7,022 & D/S &2021 \\
    DBF~\cite{DBF} & \textcolor[rgb]{0,0.5,0}{\Checkmark} & \textcolor[rgb]{1,0,0}{\XSolidBrush} & \textcolor[rgb]{1,0,0}{\XSolidBrush} &4 & 2,528  & D & 2021  \\
    Dualray~\cite{dualray} & \textcolor[rgb]{0,0.5,0}{\Checkmark} & \textcolor[rgb]{1,0,0}{\XSolidBrush} & \textcolor[rgb]{1,0,0}{\XSolidBrush} &6 & 4,371  & D & 2022 \\
    DvXray~\cite{DvXray} & \textcolor[rgb]{0,0.5,0}{\Checkmark} & \textcolor[rgb]{1,0,0}{\XSolidBrush} & \textcolor[rgb]{0,0.5,0}{\Checkmark}&15 & 5,000  & C & 2024  \\
    LDXray~\cite{ldxray} & \textcolor[rgb]{0,0.5,0}{\Checkmark} & \textcolor[rgb]{1,0,0}{\XSolidBrush} & \textcolor[rgb]{0,0.5,0}{\Checkmark}&12 & 146,997  & D & 2024  \\
    \midrule
    STCray~\cite{STING-BEE} & \textcolor[rgb]{1,0,0}{\XSolidBrush} & \textcolor[rgb]{0,0.5,0}{\Checkmark} & \textcolor[rgb]{0,0.5,0}{\Checkmark}&21 & 46,642  & D/S/ZS & 2024  \\
    PIXray Caption~\cite{PIXray-Caption} & \textcolor[rgb]{1,0,0}{\XSolidBrush} & \textcolor[rgb]{0,0.5,0}{\Checkmark} & \textcolor[rgb]{0,0.5,0}{\Checkmark}&15 & 5,046  & D/ZS & 2025  \\
    \midrule
    DualXrayBench & \textcolor[rgb]{0,0.5,0}{\Checkmark} & \textcolor[rgb]{0,0.5,0}{\Checkmark} & \textcolor[rgb]{0,0.5,0}{\Checkmark}& 12 & 45,613  & D/ZS & 2025  \\    
    \bottomrule
  \end{tabular}
    \caption{
  A comprehensive overview of X-ray datasets used in security inspection, detailing the number of categories ($N_{c}$) and images containing prohibited items ($N_{p}$). The tasks include Classification (C), Detection (D), Segmentation (S), and Zero-Shot (ZS).
  }
\label{table:2}
\end{table}

% \begin{figure*}[t]        % 关键：figure* 才能跨双栏
%   \centering
%   \includegraphics[width=\linewidth]{figures/dual-view-2.pdf}
%   \vspace{-0.3in}
%   \caption{The construction pipeline DualXrayCap}
%   \label{fig:2}
%   \vspace{-0.2in}
% \end{figure*}

\begin{figure*}[!t]       
  \centering
    \vspace{-0.2in}
  \includegraphics[width=1\linewidth]{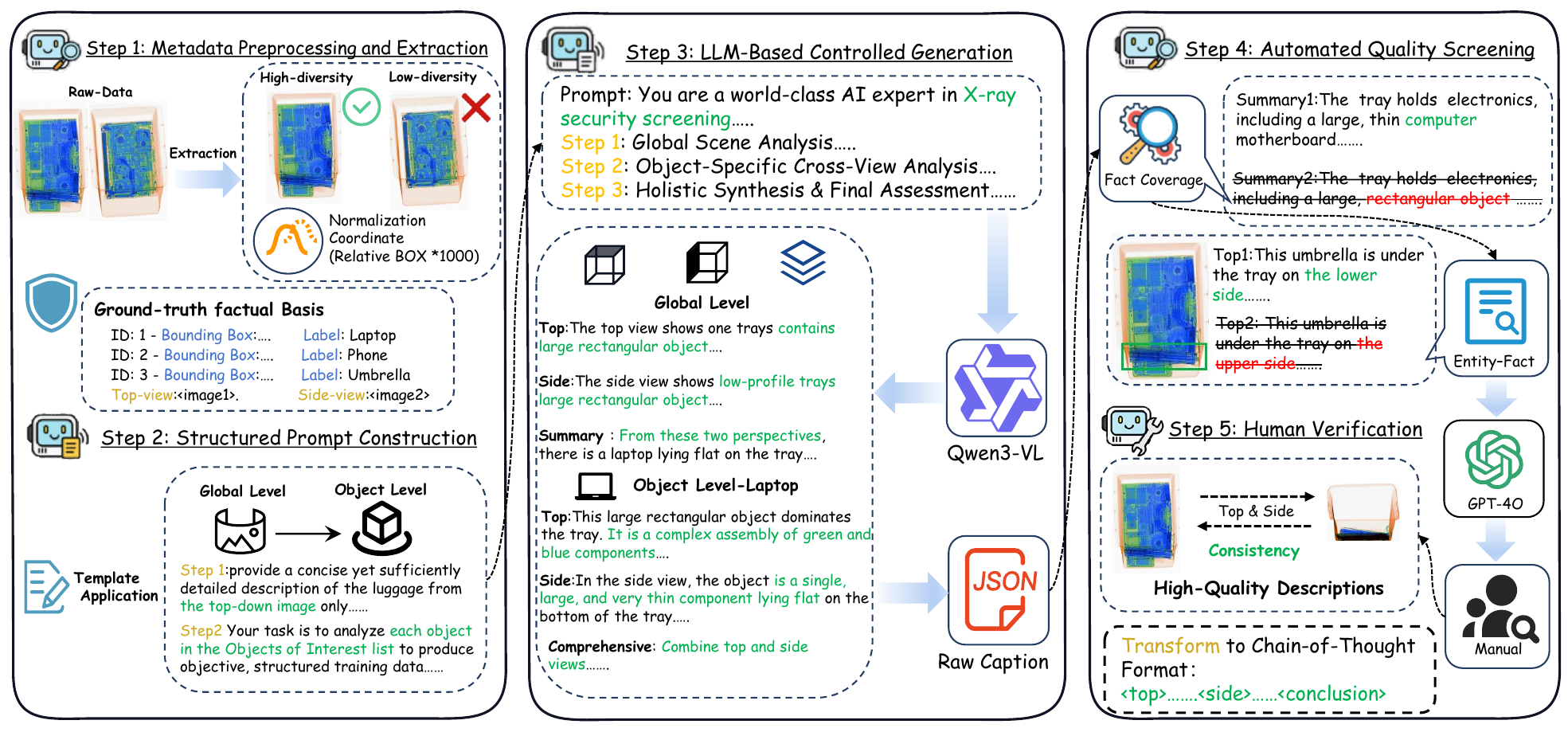}
  % \vspace{-0.8in}
  % \vspace{-20pt}
  \caption{Construction pipeline of the DualXrayBench caption corpus, illustrating metadata preprocessing and extraction, structured prompt design, LLM-based controlled caption generation, automated quality filtering, and expert verification to ensure high-quality captions for X-ray prohibited-item detection and reasoning.}
  \label{fig:3}
  \vspace{-0.15in}
\end{figure*}
\section{Related Work}
\label{sec:relate}

% Recent advances in computer-aided X-ray screening have been largely driven by deep learning and the release of public datasets such as SIXray \cite{SIXray}, PIDray \cite{pidray_iccv}, and HiXray \cite{Hixray}, which have propelled research on object detection and segmentation of prohibited items. Although these datasets comprise hundreds of thousands of scans and address challenges like class imbalance and occlusion, most adhere to a single-view, closed-set paradigm, offering limited sample diversity, sparse annotations, and insufficient representation of real-world concealment complexity. More importantly, such single-view resources cannot support multimodal architectures that rely on cross-perspective reasoning. In real security inspection, however, human operators routinely analyze top- and side-view scans to mentally reconstruct 3D structures and resolve occlusion ambiguities. This motivates dual-view research, exemplified by datasets such as DvXray \cite{dualray}, which provide paired scans but remain constrained by scale and annotation granularity. Existing dual-view models mainly employ feature fusion strategies \cite{ldxray,dagnet} that treat the auxiliary view as supplementary visual data, lacking explicit modeling of cross-perspective visibility, occlusion, and geometric consistency—core elements required for human-like spatial reasoning in dual-view X-ray analysis.

\textbf{X-ray Baggage Security Datasets.} Recent advances in computer-aided X-ray screening have been largely driven by deep learning and the release of public datasets \cite{GDXray,SIXray,pidray_iccv,OPIXray,Hixray} such as SIXray \cite{SIXray}, PIDray \cite{pidray_iccv}, and HiXray \cite{Hixray}, which have propelled research on object detection and segmentation of prohibited items, as summarized in Table~\ref{table:2}. Although these datasets comprise hundreds of thousands of scans and address challenges like class imbalance and occlusion, most adhere to a single-view, closed-set paradigm. Crucially, in real-world security inspection, human operators routinely examine both top and side view scans to mentally reconstruct 3D object structures and resolve occlusion-induced ambiguities. This motivates dual-view research \cite{mvxray,deei6,dualray,DvXray,ldxray}, exemplified by datasets such as DvXray \cite{dualray}, which provide paired scans but remain constrained by scale and annotation granularity. Existing dual-view models primarily rely on feature fusion \cite{ldxray,dagnet}, treating the auxiliary view as additional imagery without explicitly modeling cross-perspective visibility or geometric consistency key factors for human-level spatial reasoning in dual-view X-ray analysis

% Existing dual-view models mainly employ feature fusion strategies \cite{ldxray,dagnet} that treat the auxiliary view as supplementary visual data, lacking explicit modeling of cross-perspective visibility, occlusion, and geometric consistency—core elements required for human-like spatial reasoning in dual-view X-ray analysis.

\textbf{Vision-Language Models in X-ray Security.} The success of large Vision-Language Models (VLMs) \cite{li2021glip,liu2023llava,wasim2024vgdino}, exemplified by foundational models like GLIP \cite{li2021glip}, has opened new avenues for open-vocabulary detection and semantic understanding. Recent research has adapted VLMs to specialized domains \cite{llava_med}, including medical imaging and X-ray security. Efforts like OVXD \cite{ovxd} adapts CLIP for open-vocabulary X-ray detection but relies solely on word-level textual supervision, limiting its semantic understanding. OVPID introduces the PIXray-Caption dataset \cite{PIXray-Caption} with 5,046 image–caption pairs to alleviate the closed-set limitation, yet its limited caption quantity and coarse image-level descriptions constrain fine-grained visual–language alignment. STCray \cite{STING-BEE} brings instruction-following to X-ray analysis via text–image pairs but remains restricted to single-view alignment without modeling cross-view geometry. Overall, existing work advances X-ray vision–language modeling yet still lacks fine-grained data and explicit cross-view reasoning.
% STCray \cite{STING-BEE} extends X-ray analysis to the multimodal instruction-following paradigm by introducing text–image pairs, but it remains confined to single-view visual–text alignment without modeling cross-view geometric correspondence. Overall, these studies advance vision–language modeling in X-ray security but still lack fine-grained data and explicit cross-view reasoning.

% Together, these approaches highlight the growing integration of vision–language modeling in X-ray security inspection, yet they remain constrained by limited data granularity and the absence of explicit cross-view reasoning mechanisms.

% Efforts like STCray \cite{STING-BEE} and related works \cite{PIXray-Caption} have successfully introduced text captions to the X-ray domain to alleviate the closed-set limitation and enable multi-modal instruction following. Despite this progress, these approaches remain constrained by two major factors: first, the inherent domain shift between natural images (used for pre-training) and X-ray scans; and second, their reliance on single-view data. Our work moves beyond this limitation by recognizing that the auxiliary view provides a potent structural constraint that can be formalized through a structured language framework. This allows us to conceptualize and leverage the geometric information within the dual-view setup as a powerful form of structured language, which is missing in all prior VLM efforts in this field.

\begin{figure*}[t]        % 关键：figure* 才能跨双栏
  \centering
  \vspace{-0.3in}
  \includegraphics[width=1\linewidth]{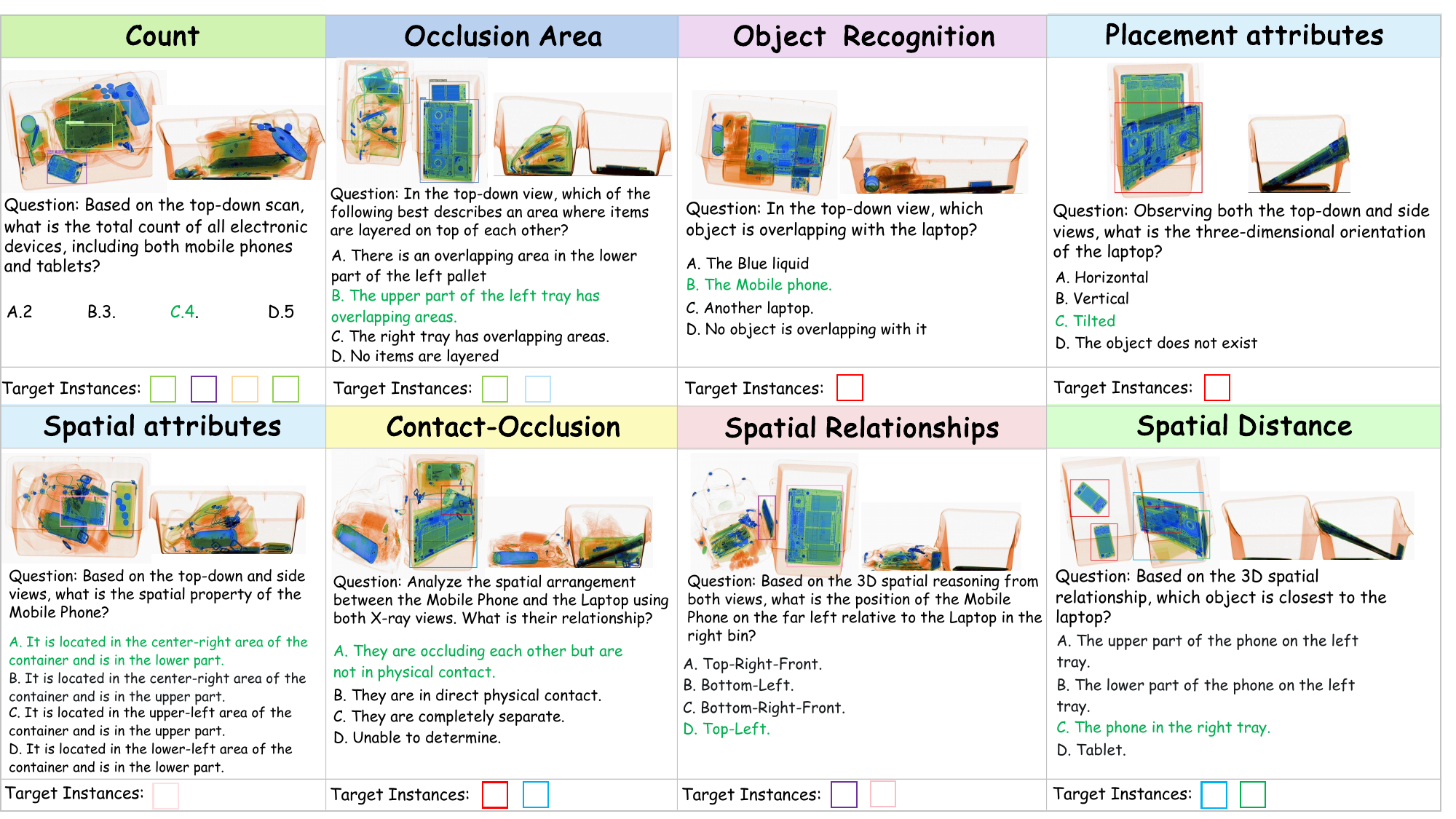}
  \vspace{-0.2in}
  % \caption{Representative examples from DualXrayBench illustrating the eight diagnostic task types. Each example visualizes paired top- and side-view images with annotated evidence and corresponding questions. Tasks cover perception (counting, recognition), relational reasoning (spatial alignment), occlusion inference (visibility and containment), and attribute understanding (posture and geometric properties), showcasing the benchmark’s comprehensive coverage of cross-perspective reasoning challenges.}
  \caption{Representative examples from DualXrayBench illustrating eight diagnostic tasks. Each example pairs top- and side-view images with annotated evidence and corresponding questions. Tasks encompass perception (counting, recognition), relational reasoning (spatial alignment), occlusion inference (visibility and containment), and attribute understanding (posture, geometry), highlighting the benchmark's coverage of cross-perspective reasoning challenges.}
  \label{fig:4}
  \vspace{-0.15in}
\end{figure*}

\section{DualXrayBench} 
\label{sec:DualXrayCap and DualXrayBench}

To enable structured cross-perspective visual–language reasoning, we introduce \textbf{DualXrayBench}, including a large-scale hierarchical caption corpus built upon the dual-view LDXray dataset~\cite{ldxray}. LDXray contains 146,997 paired top and side view X-ray images with annotated bounding boxes and categories. Based on these, we generate 45,613 verified dual-view caption pairs capturing both scene-level and fine-grained cross-view correspondences across 12 classes: Mobile Phone (MP), Orange Liquid (OL), Portable Charger 1, lithium-ion prismatic cell (PC1), Portable Charger 2, lithium-ion cylindrical cell (PC2), Laptop (LA), Green Liquid (GL), Tablet (TA), Blue Liquid (BL), Columnar Orange Liquid (CO), Nonmetallic Lighter (NL), Umbrella (UM) and Columnar Green Liquid (CG).

\subsection{Corpus Construction Pipeline}

\textbf{Pipeline Overview.} As illustrated in Fig.~\ref{fig:3}, the caption corpus is built through a semi-automated, five-stage pipeline ensuring factual correctness, linguistic coherence, and cross-view consistency.

\textbf{(1) Preprocessing.}
We filter low-diversity samples and normalize object bounding boxes to a $[0,1000]$ scale, producing resolution-independent structured metadata comprising category and spatial information.

\textbf{(2) Structured Prompting.}
Each image pair and its metadata are formatted into a hierarchical prompt instructing the model to describe the top and side view scenes, provide object-level analyses, and summarize global semantics under strict JSON constraints to avoid hallucination.

\textbf{(3) LLM-Based Generation.}
Using Qwen3-VL-235B-A22B-Thinking \cite{qwen3vl}, we generate hierarchical captions including global layout, per-object reasoning, and holistic summaries. Descriptions explicitly encode complementary cross-view cues (e.g., “flat in top-view but vertically extended in side-view”).

\textbf{(4) Automated Screening.} A three-stage validation ensures quality through: (i) fact coverage, (ii) entity–fact alignment with bounding boxes, and (iii) cross-view consistency checking via GPT-4o \cite{gpt4o}. Samples failing coverage ($<80\%$), alignment ($<50\%$), and consistency ($>50\%$ contradiction) are discarded.

\textbf{(5) Human Verification.} Domain experts conduct multi-round review to ensure spatial and semantic accuracy, revising inconsistent or ambiguous cases. The verified corpus exhibits high factual precision and naturalness.

\textbf{Resulting Corpus.} The final Corpus provides rich, hierarchical dual-view captions suitable for visual–language reasoning and instruction tuning. Furthermore, its structured form supports the construction of the \textbf{GSXray dataset} by converting captions into a Chain-of-Thought format $\text{\textless top\textgreater,\textless side\textgreater,\textless conclusion\textgreater}$ with the assistance of GPT-4o \cite{gpt4o}. This design explicitly supervises cross-perspective reasoning alignment in multimodal models. \textbf{Details are provided in the Supplementary Materials.}

\subsubsection{Data Statistics}

\textbf{Category and Instance Distributions.} DualXrayCap contains 45,613 paired images with a total of 139,956 instances spanning 12 common categories, as summarized in Table \ref{dataset info}. The dataset exhibits diverse annotation patterns, with 5,211 images containing a single instance and 38,967 images having multiple instances, resulting in an average of 3.07 annotations per image. Moreover, 22,772 images contain overlapping objects with an Intersection over Minimum (IoM) greater than 0.3, indicating frequent occlusions and complex spatial arrangements.

\begin{table}[!h]
\setlength{\tabcolsep}{0.4mm}
  \label{dataset info}
  % \vspace{-0.2in}
  \fontsize{5.5}{11.8}\selectfont
  \centering
\scalebox{1.1}{
  \begin{tabular}{l|cccccccccccc|c}
    \toprule
    Category & MP & OL & PC1 & PC2 & LA & GL & TA & BL & CO & NL & UM & CG & Total \\
    \midrule
    Number & 59,003  & 18,684  & 9,998  & 17,274  & 13,453  & 11,482  & 6,796  & 1,677  & 513  & 487  & 298  & 296& 139,956 \\
    \bottomrule
  \end{tabular}
  }
  \caption{Category-wise statistics of DualXray corpus.}
  \label{dataset info}
  \vspace{-0.1in}
\end{table}

\begin{figure*}[!t]        
  \centering
  \vspace{-0.2in}
  \includegraphics[width=1\linewidth]{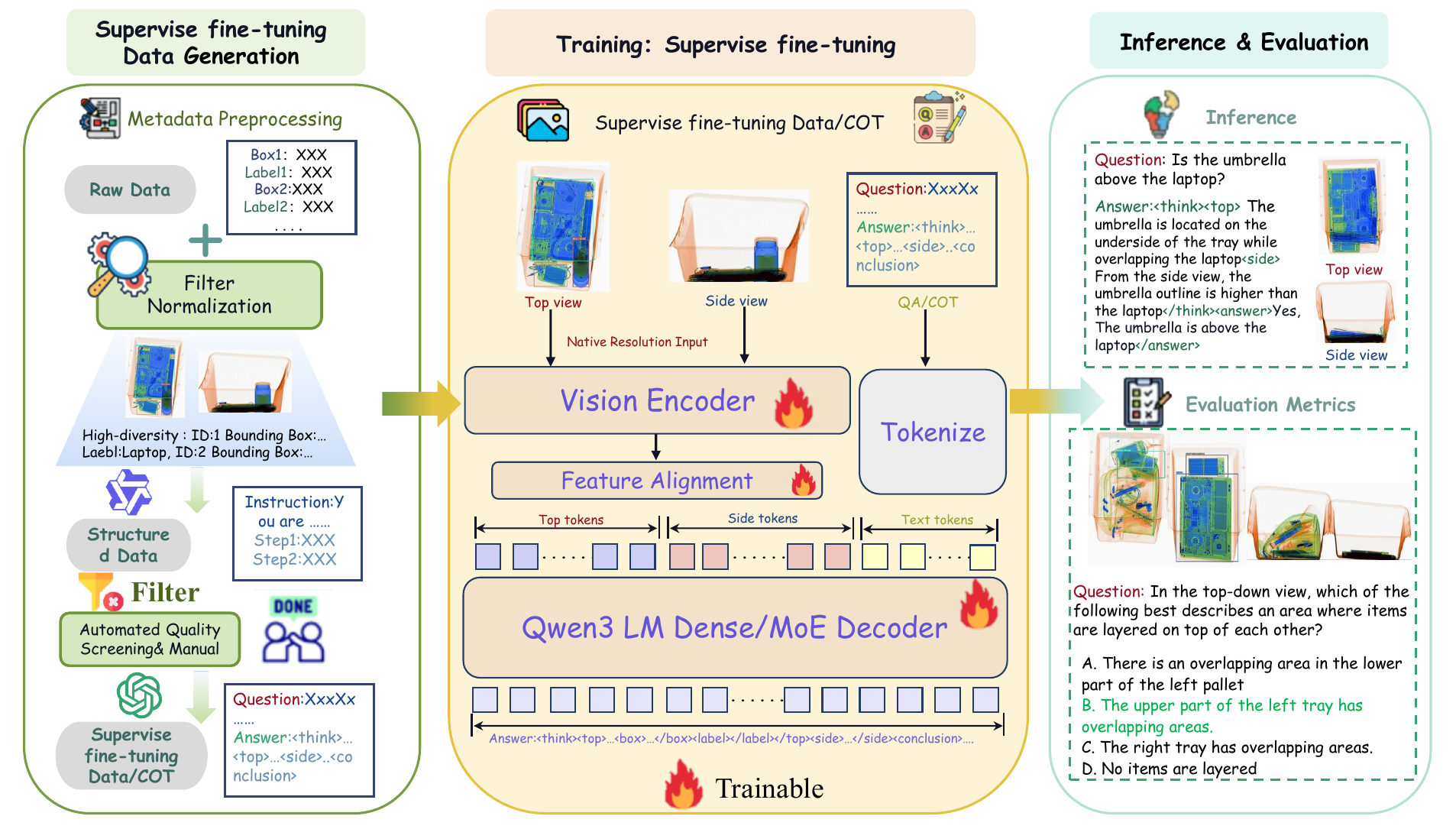}
  % \vspace{-0.3in}
  % \caption{Overall Architecture of the DualXrayBench–GSR Framework.(1) A structured and high-diversity data generation pipeline that constructs DualXrayBench and GSXray CoT supervision from raw multi-view X-ray metadata;(2) A supervised fine-tuning framework that jointly learns dual-view geometric correspondences and cross-modal semantics, treating side-view imagery as a language-like modality; and (3) A unified inference and evaluation protocol for cross-view reasoning, consistency assessment, and spatial understanding across eight diagnostic tasks.}
  \caption{Overall architecture of the DualXrayBench–GSR framework: (1) A structured data generation pipeline for creating DualXrayBench and GSXray CoT supervision from raw multi-view X-ray metadata; (2) A supervised fine-tuning framework that learns dual-view geometric correspondences and cross-modal semantics, treating side-view imagery as a language-like modality; (3) A unified inference and evaluation protocol for cross-view reasoning, consistency assessment, and spatial understanding across eight diagnostic tasks.}
  \label{fig:5}
  \vspace{-0.15in}
\end{figure*}

% Unlike conventional single-view evaluation protocols, DualXrayBench adopts a dual-view assessment paradigm that explicitly measures models’ reasoning ability, interpretability, cross-view consistency, and spatial understanding across eight diagnostic sub-tasks.

\subsection{DualXrayBench Construction Pipeline}
We introduce \textbf{DualXrayBench}, a multi-task benchmark comprising 1,594 expert-verified visual question–answer pairs. It is constructed through a semi-automated pipeline that converts structured image–text data from the DualXray corpus into a comprehensive suite of measurable reasoning tasks, with strict checks to prevent overlap with the GSXray dataset. The benchmark is produced via a three-stage process that integrates automated generation with expert validation to ensure reliability in cross-perspective visual–language reasoning.

% We introduce \textbf{DualXrayBench}, a multi-task benchmark of 1,594 expert-verified visual question–answer pairs. Built through a semi-automated, three-stage pipeline, it converts structured image–text annotations from the DualXray corpus into measurable reasoning tasks, while ensuring zero overlap with GSXray. Expert validation is incorporated to guarantee reliable evaluation of cross-perspective visual–language reasoning.

\textbf{Stage I Source corpus preparation}
We randomly sample 5,000 instances from the DualXray corpus, which provides top and side view scene analyses, fine-grained object-level descriptions, and holistic summaries. From these, we extract structured scene facts including object categories, spatial relationships, and occlusion and containment relations to serve as verified inputs for question generation.

\textbf{Stage II Multi-task Q\&A generation}
Each sample is used to generate four distinct question–answer pairs via controlled LLM prompting, corresponding to: (1) \emph{Perceptual} (counting/identification), (2) \emph{Relational} (spatial reasoning), (3) \emph{Occlusion} (visibility inference) and (4) \emph{Attribute} (pose/property recognition). Each includes a reference answer and traceable evidence (bounding boxes and labels).

\textbf{Stage III Human review \& difficulty calibration}
All samples undergo two-stage review: expert quality control for clarity and answerability, followed by model-consensus filtering to remove trivial or overly hard cases. The final benchmark contains 1,594 high-quality, human-verified samples, ensuring both difficulty balance and traceable evaluation integrity.

\subsubsection{Task Definition}
To comprehensively evaluate cross-perspective reasoning, \textbf{DualXrayBench} defines four task families \emph{Perceptual}, \emph{Relational}, \emph{Occlusion}, and \emph{Attribute} each comprising representative sub-tasks that emphasize dual-view understanding (Fig.~\ref{fig:4}).

\textbf{Perceptual Tasks.}
This category includes \emph{Counting} (CT) and \emph{Object Recognition} (OR). Unlike conventional single-view setups, both tasks are constructed under occlusion conditions to highlight the complementary role of side-view information. CT evaluates a model’s ability to count partially hidden instances, while OR examines cross-view category identification consistency.

\textbf{Relational Tasks.}
We design \emph{Spatial Relation} (SR) and \emph{Spatial Distance} (SD) tasks to assess geometric reasoning across perspectives. SR requires inferring the relative position of object pairs (e.g., above/below, front/behind), while SD focuses on estimating inter-object distances, particularly along the depth (\(z\))-axis challenging for single view perception.

\textbf{Occlusion Tasks.}
This family comprises \emph{Occluded Area Recognition} (OA) and \emph{Contact–Occlusion Judgment} (CO). OA targets identifying occluded regions by integrating complementary visibility cues across views, while CO requires reasoning about whether two objects are physically contacting or mutually occluding demanding explicit cross-view inference.

\textbf{Attribute Tasks.}
Finally, \emph{Placement Attribute} (PA) and \emph{Spatial Attribute} (SA) tasks evaluate the model’s understanding of object posture and spatial configuration. The former focuses on pose level properties (e.g., standing, flat, tilted), and the latter examines localization in 3D space, especially along the vertical dimension.

Together, these eight sub-tasks form a structured, multi-faceted evaluation protocol that probes not only recognition accuracy but also the depth of geometric and causal reasoning enabled by dual-view perception, with the number of instances per sub-task summarized in the Table \ref{dualXrayBench info}.

\begin{table}[!h]
  \centering
  \small
  \setlength{\tabcolsep}{1.2mm}
  % \fontsize{5.5}{11.8}\selectfont
 \begin{tabular}{l|cccccccc|c}
  % \begin{tabularx}{\linewidth}{l *{8}{>{\centering\arraybackslash}X} c}
    \toprule
    Task &  \textbf{CT} &  \textbf{OR} &  \textbf{SR} &  \textbf{SD} &  \textbf{OA} &  \textbf{CO} &  \textbf{PA} &  \textbf{SA} & Total \\
    \midrule
    Number & 200 & 200 & 200 & 200 & 196 & 199 & 199 & 200 & 1594 \\
    \bottomrule
  \end{tabular}
\caption{Sample distribution of the eight diagnostic sub-tasks.}
\label{dualXrayBench info}
\vspace{-0.2in}
\end{table}

% \begin{table}[!h]
% \setlength{\tabcolsep}{0.46mm}
%   \fontsize{5.5}{11.8}\selectfont
%   \centering
%   \begin{tabular}{l|cc|cc|cc|cc|c}
%     \toprule
%     Task & CT & OR & SR & SD & OA & CO & PA & SA & Total \\
%     \midrule
%     Number & 200  & 200  & 200  & 200  & 196  & 199  & 199  & 200  & 1594 \\
%     \bottomrule
%   \end{tabular}
%   \caption{The number of instances in the training and testing sets.}
%   \label{dualXrayBench info}
%   \vspace{-0.1in}
% \end{table}

\section{Geometric-Semantic Reasoner (GSR)} 
\label{sec:Geometric-Semantic Reasoner}
Building upon these data, we propose the GSR, a multimodal model that jointly learns textual semantics and dual-view geometric correspondences, treating the second-view image as a structured ``language-like modality".
\subsection{GSXray dataset}

To enable supervision for cross-view geometric semantic reasoning, we construct the  \textbf{G}eometric–\textbf{S}emantic dual-view X-ray dataset (\textbf{GSXray}), which comprises 44,019 dual-view QA samples organized into structured Chain-of-Thought sequences \texttt{<top>}, \texttt{<side>}, and \texttt{<conclusion>}. Unlike conventional SFT CoT datasets, the \textbf{GSXray} designed for dual view X-ray inspection decomposes reasoning into view specific observations that are fused into a unified semantic conclusion. This structure offers fine-grained supervision for modeling the shift from geometric perception to high-level semantics, akin to how human inspectors integrate top side views to resolve occlusions. These CoT sequences are automatically transformed from  DualXray Corpus with lightweight GPT assistance and minimal manual oversight. Representative examples are illustrated in the figure \ref{fig:5}. \textbf{More details are provided in the Supplementary Materials.}

\subsection{GSR Implementation}
To enable fine-grained cross-perspective understanding in X-ray imagery, we propose the \textbf{Geometric–Semantic Reasoner (GSR)} a multimodal reasoning architecture built upon the Qwen3-VL-MoE-8B \cite{qwen3vl} foundation model. GSR extends the visual–language alignment capability of large multimodal models (LMMs) by explicitly encoding geometric correspondence and semantic hierarchy across dual views. As shown in figure \ref{fig:5}, the architecture comprises three key components: a vision encoder \(\bm{E}\), a Feature Alignment \(\bm{A}\), and a language reasoning model \(\bm{L}\).

The vision encoder $\bm{E}$ follows the ViT-L/14 design within the Qwen3-VL-MoE framework. It independently processes the paired dual-view X-ray inputs top view $\bm{x}_{\text{top}}$ and side-view $\bm{x}_{\text{side}}$ through a shared-weight transformer backbone equipped with 3D convolutional patch embedding and rotary positional encoding. The encoder extracts dense feature representations that preserve both geometric structures and spatial context, producing:

\begin{equation}
\bm{f}_{i} = \bm{E}(\bm{x}_{i}) \in \mathbb{R}^{m \times n},
\end{equation}
where $\bm{f}_{i}$ denotes a sequence of vision tokens encoding object-level and contextual cues from the top and side view image. The multi-level attention mechanism within the MoE transformer allows for efficient feature aggregation across scales, capturing view-dependent relationships essential for X-ray scene interpretation.

A lightweight multimodal alignment module $\bm{A}$ implemented as a two-layer MLP with GeLU activation maps visual embeddings into the LLM's semantic space:

\begin{equation}
\bm{f}_{i}' = \bm{A}(\bm{f}_{i}).
\end{equation}
This alignment bridges the geometric representation space of the vision encoder and the linguistic embedding space of the LLM, producing the projected feature tokens $\bm{f}_{i}' \in \mathbb{R}^{k}$. These tokens are then fed into the LLM as structured multimodal inputs.

% -----------------------------------------------

\begin{table*}[!t]
 \centering
 \vspace{-0.2in}
 \Large
 \resizebox{\linewidth}{!}{
    \begin{tabular}{l|ccc|cc|cc|cc|cc}
    \toprule
    \multirow{2}{*}{Method}& \multicolumn{3}{c|}{Overall}& \multicolumn{2}{c|}{Perception}& \multicolumn{2}{c|}{Relational}& \multicolumn{2}{c|}{Occlusion}& \multicolumn{2}{c}{Attribute} \\
    & Acc &F1 & mIOU & CT & OR & SR & SD & OA & CO & PA & SA \\ 
    \midrule
    GPT-4o~\cite{gpt4o} & 47.0 &49.2 &16.5 &46.5 & 45.5 & 43.0 & 38.5 & 43.4 & 57.8 & 47.5 & 59.8  \\
    GPT-o3~\cite{o3} & 49.3 &52.6 &19.7 &47.1 & 48.9 & 50.5 & 36.4 & 43.5 & 59.2 & 52.5 & 56.3  \\
    Gemini-2.5-Flash~\cite{gemini-2.5-flash} & 45.1 &48.9 &14.7 &43.0 & 42.9 & 41.2 & 34.5 & 42.8 & 56.0 & 51.0 & 49.4  \\
    Gemini-2.5-Pro~\cite{gemini-2.5-pro} & 58.6 &60.5 &28.7 &56.3 & 43.0 & 44.3 & 61.5 & 52.9 & 74.6 & 64.3 & 71.9  \\
    \hline
    Qwen2.5-VL-7B~\cite{Qwen2.5-VL} & 50.3 &53.0 & 20.3 & 35.5 & 43.5 & 47.5 & 48.5 & 43.4 & 52.8 &  65.3 & 65.8  \\
    Qwen2.5-VL-32B~\cite{Qwen2.5-VL} & 53.0 & 55.4 & 24.5 & 45.0 & 42.5 & 47.0 & 54.0 & 50.0 & 53.8 & 58.1 & 74.0 \\
    Qwen2.5-VL-72B~\cite{Qwen2.5-VL} & 57.1 & 59.5 & 26.3 & 54.0 & 43.7 & 39.5 & 59.0 & 52.6 & 72.4 & 65.6 & 70.5  \\
    Qwen3-VL-4B~\cite{qwen3vl} & 47.2 & 50.2 & 20.2 & 18.2 & 41.5 & 45.0 & 46.5 & 46.9 & 53.3 & 52.9 & 59.5 \\
    Qwen3-VL-8B~\cite{qwen3vl} & 53.5 & 56.6 & 25.4 & 45.5 & 42.9 & 38.0 & 47.0 & 54.6 & 71.9 & 62.6 & 66.5 \\
    Qwen3-VL-30B~\cite{qwen3vl} & 51.8 & 53.3 & 25.4 & 49.2 & 47.0 & 46.0 & 54.5 & 50.5 & 51.3 & 53.3 & 62.5 \\
    Qwen3-VL-235B~\cite{qwen3vl} & 58.8 & 65.5 & 26.0 & 54.0 & 55.8 & 40.5 & \textbf{61.0} & 56.6 & 69.3 & 54.0 & \textbf{79.0} \\
    InternVL2-8B~\cite{chen2024internvl} & 33.8 & 37.4 & 8.4 & 29.9 & 35.5 & 29.9 & 31.6 & 30.8 & 39.2 & 36.7 & 36.4 \\ 
    InternVL2.5-8B~\cite{chen2024internvl} & 36.3 & 40.6 & 10.2 & 27.3 & 36.9 & 30.8 & 32.5 & 38.8 & 43.9 & 43.3 & 37.2 \\
    InternVL3-8B~\cite{zhu2025internvl3} & 43.6 & 45.4 & 15.3 & 34.3 & 32.1 & 28.6 & 34.8 & 44.0 & 57.2 & 56.8 & 61.0  \\  
    InternVL3-30B~\cite{zhu2025internvl3} & 45.4 & 47.3 & 15.7 & 32.5 & 40.1 & 29.7 & 36.5 & 46.0 & 58.4 & 60.3 & 59.7  \\ 
    InternVL3-78B~\cite{zhu2025internvl3} & 48.9 & 51.6 & 16.4 & 43.5 & 39.1 & 28.5 & 41.2 & 48.0 & 65.8 & 63.9 & 61.2 \\ 
    InternVL3.5-8B~\cite{wang2025internvl3_5} & 45.4 & 51.5 & 18.1 & 43.7 & 29.5 & 28.0 & 36.5 & 49.0 & 61.8 & 57.6 & 58.0 \\ 
    InternVL3.5-30B~\cite{wang2025internvl3_5} & 46.1 & 49.3 &  18.5 & 37.5 & 41.5 & 30.5 & 37.5 & 46.4 & 60.9 & 61.8 & 54.0 \\
    InternVL3.5-241B~\cite{wang2025internvl3_5} & 51.3 & 53.8 & 22.0 & 46.8 & 43.0 & 30.2 & 42.5 & 49.6 & 73.8 & \textbf{66.2} & 58.3 \\
    LLaVA1.5-7B~\cite{liu2023llava} & 15.4 & 20.5 & 6.2 & 19.5 & 11.0 & 16.0 & 9.5 & 15.3 & 21.6 & 12.6 & 18.0  \\
    LLaVA-Next-7B~\cite{liu2024llavanext} & 24.5 & 28.3 & 13.1 & 20.6 & 27.1 & 24.0 & 23.1 & 20.2 & 27.3 & 27.8 & 25.4  \\
    LLaVA-Next-34B~\cite{liu2024llavanext} & 28.3 & 36.5 & 14.5 & 24.0 & 31.7 & 26.0 & 26.3 & 24.7 & 38.2 & 28.1 & 27.4  \\ 
    LLaVA-Next-110B~\cite{liu2024llavanext} & 33.6 & 38.8 & 16.8 & 27.0 & 25.1 & 24.5 & 31.5 & 31.8 & 53.0 & 40.0 & 36.0  \\
    LLaVA-OneVision-7B~\cite{liu2023improvedllava} & 32.3 & 36.8 & 15.6 & 34.5 & 33.0 & 28.5 & 33.7 & 29.1 & 30.8 & 31.5 & 37.0  \\
    LLaVA-OneVision-72B~\cite{liu2023improvedllava} & 42.2 & 45.1 & 19.2 & 34.5 & 40.0 & 40.0 & 41.5 & 44.9 & 45.2 & 47.7 & 43.7  \\   
    \midrule
     STING-BEE~\cite{liu2023improvedllava} & 23.8 & 29.8 & 13.2 & 22.6 & 26.4 & 23.6 & 24.1 & 19.4 & 23.4 & 24.5& 26.8 \\
    \midrule
    \rowcolor{ours} Qwen2.5-VL-7B(Ours) &55.1 $_{\color[rgb]{0.2,0.7,0.2}{\uparrow 4.8\%}}$ &62.2 $_{\color[rgb]{0.2,0.7,0.2}{\uparrow 9.2\%}}$&28.7 $_{\color[rgb]{0.2,0.7,0.2}{\uparrow 8.4\%}}$&46.2 $_{\color[rgb]{0.2,0.7,0.2}{\uparrow 10.7\%}}$ &45.3 $_{\color[rgb]{0.2,0.7,0.2}{\uparrow 1.8\%}}$&47.9 $_{\color[rgb]{0.2,0.7,0.2}{\uparrow 0.4\%}}$& 55.7$_{\color[rgb]{0.2,0.7,0.2}{\uparrow 7.2\%}}$&51.6 $_{\color[rgb]{0.2,0.7,0.2}{\uparrow 8.2\%}}$&64.5 $_{\color[rgb]{0.2,0.7,0.2}{\uparrow 11.7\%}}$&66.1$_{\color[rgb]{0.2,0.7,0.2}{\uparrow 0.8\%}}$&63.5 \\
    \rowcolor{ours} InternVL3.5-8B(Ours) &50.3 $_{\color[rgb]{0.2,0.7,0.2}{\uparrow 4.9\%}}$ &54.5 $_{\color[rgb]{0.2,0.7,0.2}{\uparrow 3.0\%}}$&23.2 $_{\color[rgb]{0.2,0.7,0.2}{\uparrow 5.1\%}}$&52.0 $_{\color[rgb]{0.2,0.7,0.2}{\uparrow 8.3\%}}$ &44.0 $_{\color[rgb]{0.2,0.7,0.2}{\uparrow 14.5\%}}$&24.5 &44.0 $_{\color[rgb]{0.2,0.7,0.2}{\uparrow 7.5\%}}$&52.0 $_{\color[rgb]{0.2,0.7,0.2}{\uparrow 3\%}}$&73.9 $_{\color[rgb]{0.2,0.7,0.2}{\uparrow 12.1\%}}$&52.3&60.0 $_{\color[rgb]{0.2,0.7,0.2}{\uparrow 2.0\%}}$ \\
    \rowcolor{ours} LLaVA-Next-7B(Ours) &29.5 $_{\color[rgb]{0.2,0.7,0.2}{\uparrow 5.0\%}}$ &34.0 $_{\color[rgb]{0.2,0.7,0.2}{\uparrow 5.7\%}}$ &23.1 $_{\color[rgb]{0.2,0.7,0.2}{\uparrow 10.0\%}}$ & 19.0  &37.6 $_{\color[rgb]{0.2,0.7,0.2}{\uparrow 10.5\%}}$ & \textbf{54.1}$_{\color[rgb]{0.2,0.7,0.2}{\uparrow 1.1\%}}$ &33.2 $_{\color[rgb]{0.2,0.7,0.2}{\uparrow 10.1\%}}$ &26.2 $_{\color[rgb]{0.2,0.7,0.2}{\uparrow 6.0\%}}$&28.9 $_{\color[rgb]{0.2,0.7,0.2}{\uparrow 1.6\%}}$ & 26.6 &38.8 $_{\color[rgb]{0.2,0.7,0.2}{\uparrow 13.4\%}}$ \\
    \rowcolor{ours} LLaVA-OneVision-7B(Ours) &39.3 $_{\color[rgb]{0.2,0.7,0.2}{\uparrow 7.0\%}}$ &47.2 $_{\color[rgb]{0.2,0.7,0.2}{\uparrow 10.4\%}}$&28.3 $_{\color[rgb]{0.2,0.7,0.2}{\uparrow 12.7\%}}$&38.2 $_{\color[rgb]{0.2,0.7,0.2}{\uparrow 3.7\%}}$ &37.6 $_{\color[rgb]{0.2,0.7,0.2}{\uparrow 4.6\%}}$&30.1$_{\color[rgb]{0.2,0.7,0.2}{\uparrow 1.6\%}}$ &42.6 $_{\color[rgb]{0.2,0.7,0.2}{\uparrow 8.9\%}}$&37.3 $_{\color[rgb]{0.2,0.7,0.2}{\uparrow 8.2\%}}$&45.7 $_{\color[rgb]{0.2,0.7,0.2}{\uparrow 14.9\%}}$&43.1$_{\color[rgb]{0.2,0.7,0.2}{\uparrow 11.6\%}}$&39.8 $_{\color[rgb]{0.2,0.7,0.2}{\uparrow 2.8\%}}$ \\
    \rowcolor{ours} \textbf{GSR-8B (Ours)} & \textbf{65.4}$_{\color[rgb]{0.2,0.7,0.2}{\uparrow 11.9\%}}$ & \textbf{70.6}$_{\color[rgb]{0.2,0.7,0.2}{\uparrow 14\%}}$ & \textbf{52.3} $_{\color[rgb]{0.2,0.7,0.2}{\uparrow 26.9\%}}$& \textbf{63.5}$_{\color[rgb]{0.2,0.7,0.2}{\uparrow 18\%}}$ & \textbf{65.8} $_{\color[rgb]{0.2,0.7,0.2}{\uparrow 22.9\%}}$& 44.0 $_{\color[rgb]{0.2,0.7,0.2}{\uparrow 6\%}}$& \textbf{61.0}$_{\color[rgb]{0.2,0.7,0.2}{\uparrow 14\%}}$ & \textbf{67.9} $_{\color[rgb]{0.2,0.7,0.2}{\uparrow 13.3\%}}$& \textbf{80.4} $_{\color[rgb]{0.2,0.7,0.2}{\uparrow 8.5\%}}$& 64.4$_{\color[rgb]{0.2,0.7,0.2}{\uparrow 1.8\%}}$& 76.5 $_{\color[rgb]{0.2,0.7,0.2}{\uparrow 10\%}}$\\
    \bottomrule
    \end{tabular}
  }
    \caption{
    Comparison with state-of-the-art alternatives on DualXrayBench.
    All results are self-collected and each row's \textbf{Overall} equals the arithmetic mean of the eight task columns (CT, OR, SR, SD, OA, CO, PA, SA).
    The best performance is highlighted in \textbf{bold}.
    }
    \vspace{-0.1in}
    \label{tab:bench}
\end{table*}

\textbf{Hierarchical Reasoning Tokens.}
Unlike previous architectures \cite{Qwen2.5-VL,zhu2025internvl3,li2024llavaov} that treat visual embeddings as untyped tokens, GSR introduces \textbf{hierarchical reasoning tokens} \texttt{<top>}, \texttt{<side>} and \texttt{<conclusion>} to explicitly encode \textbf{cross-view semantics} \cite{STING-BEE,minigptv2, llava_med, geochat}.  
The \texttt{<top>} and \texttt{<side>} tokens distinguish visual evidence originating from orthogonal viewpoints, guiding the model to maintain spatial awareness and disambiguate occluded structures. The \texttt{<conclusion>} token serves as an aggregation signal, prompting the LLM to integrate geometric and semantic cues into a unified reasoning output.  
These tokens enable the model to dynamically adjust its reasoning granularity—from fine-grained threat localization to scene-level compositional inference—depending on the user query and the structural hierarchy implied by the token context.

\textbf{Language Reasoning.}
The \textbf{language reasoning module} $\bm{L}$, instantiated by the Qwen3-VL-MoE text decoder, jointly processes the projected visual tokens $\bm{f}_{i}'$ and the textual query sequence $\bm{t}$. Through multi-head attention and mixture-of-experts routing, the model generates responses conditioned on both the dual-view geometric cues and task-specific semantics. This process can be summarized as:
{\small
\begin{equation}
\bm{y} = \bm{L}\big([\texttt{<top>}~\bm{f}_{\text{top}}',~
\texttt{<side>}~\bm{f}_{\text{side}}',~
\texttt{<conclusion>}~\bm{t}]\big),
\end{equation}
}
where $\bm{y}$ denotes the generated reasoning output.

By explicitly incorporating geometric semantic reasoning tags, GSR overcomes a key limitation of prior multimodal models—the inability to differentiate tasks requiring distinct spatial granularity. The structured dual-view representation allows GSR to perform coherent cross-view reasoning, supporting diverse tasks such as cross-view correspondence analysis, spatial relationship inference, and multi-view consistency verification. 
% Empirically, this design leads to more interpretable and spatially consistent responses, demonstrating its effectiveness in complex security screening scenarios.

% \begin{table}[!t]
%     \centering
%     \caption{Visual Question Answering (VQA) performance across seven question categories from in-domain and cross-domain datasets.}
%     \vspace{-1em}
%     \setlength{\tabcolsep}{4pt}
%     \scalebox{0.75}[0.75]{
%     \begin{tabular}{@{}lccccccccc@{}}
%         \toprule
%         \rowcolor{Gray}
%         Model & Instance Location & Complex Reasoning & Instance Identity & Instance Counting & Misleading & Instance Attribute & Instance Interaction & Overall \\
%         \midrule
%         Florence-2 \cite{florenceV2} 
%             & 30.11 & 37.50 & 39.84 & 29.95 & 21.16 & 35.80 & 29.12 & 32.27 \\
%         MiniGPT \cite{minigptv2} 
%             & 43.22 & 60.21 & 54.79 & 28.16 & 19.64 & 32.85 & 29.74 & 36.62 \\
%         LLaVA 1.5 \cite{Llava15}  
%             & 29.73 & 56.67 & 74.04 & 34.24 & 13.76 & 40.85 & 24.03 & 41.94 \\
%         STING-BEE \cite{STING-BEE}
%             & 49.22 & 79.21 & 80.04 & 45.24 & 27.76 & 52.85 & 35.03 & 52.81 \\
%         \midrule
%         \textbf{GSR-8B (Ours)} 
%             & \textbf{52.5} & \textbf{83.0} & \textbf{83.5} & \textbf{47.0} & \textbf{29.5} & \textbf{57.0} & \textbf{41.1} & \textbf{55.3} \\
%         \bottomrule
%     \end{tabular}
% }
%     \vspace{-1em}
% \label{tab:vqa_results}
% \end{table}

\section{Experiments} 
\label{sec:Experiments}
\textbf{Implementation.}
We initialize our Geometric–Semantic Reasoner (GSR) from the Qwen3-VL-8B\cite{qwen3vl} foundation checkpoint. Both the visual encoder and the multimodal alignment module are fully unfrozen during training to enable end-to-end optimization. Fine-tuning is performed using LLaMA-Factory with mixed-precision (bfloat16) on eight NVIDIA H200 GPUs. We employ the AdamW optimizer with a base learning rate of 1e-6, cosine decay scheduling and a warmup ratio of 0.1. The global batch size is set to 256, and the model is trained for two epochs.

% resulting in a total training runtime of approximately 4.8 hours.
\subsection{DualXrayBench Evaluation}

\noindent\textbf{Evaluation Metrics.}
We evaluate model performance across the eight dual-view reasoning tasks using three metrics: mean Intersection-over-Union (mIoU), Accuracy (Acc), and F1-score. mIoU assesses spatial correspondence, Accuracy measures prediction correctness and F1-score captures the balance between precision and recall, jointly reflecting the model's geometric and semantic reasoning ability.

\noindent\textbf{Main Results.}  
We comprehensively evaluate state-of-the-art VLMs on \textbf{DualXrayBench}, covering both proprietary and open-source systems. As shown in Table~\ref{tab:bench}, most off-the-shelf models struggle with dual-view X-ray understanding, with overall accuracy typically below 50. For example, GPT-4o~\cite{gpt4o} achieves only 47.0 accuracy, and InternVL3.5-8B~\cite{wang2025internvl3_5} reaches 45.4, reflecting their limited generalization in this domain. Notably, STING-BEE, a single-view X-ray VLM, performs even worse at 23.8, underscoring its inability to handle dual-view reasoning. In contrast, our \textbf{GSR-8B} establishes a new state-of-the-art with 65.4 accuracy and 52.3 mIoU, outperforming all baselines by a large margin. It consistently leads across perception and reasoning subtasks, such as 63.5 in CT and 61.0 in SD.
% underscoring the strong cross-view reasoning and spatial grounding capabilities induced by our \textbf{GSXray} supervision.

\begin{table}[!t]
\centering
\resizebox{0.98\columnwidth}{!}{%
\begin{tabular}{cc|cccc|ccc}
\toprule

\multicolumn{2}{c|}{Model}& \textbf{Top view} & \textbf{$<$Top$>$} & \textbf{Side view} &  \textbf{$<$Side$>$}& \textbf{Acc} & \textbf{F1} & \textbf{mIoU} \\
\midrule
\multicolumn{2}{c|}{Qwen3vl-8B~\cite{qwen3vl}}&\textcolor[rgb]{1,0,0}{\XSolidBrush}&\textcolor[rgb]{1,0,0}{\XSolidBrush} &\textcolor[rgb]{1,0,0}{\XSolidBrush} & \textcolor[rgb]{1,0,0}{\XSolidBrush} & 53.5 & 56.6 & 25.4\\
\midrule

\multirow{6}{*}{
    \plusours
  }%
& & \textcolor[rgb]{0,0.5,0}{\Checkmark} &\textcolor[rgb]{1,0,0}{\XSolidBrush} &  \textcolor[rgb]{1,0,0}{\XSolidBrush} &\textcolor[rgb]{1,0,0}{\XSolidBrush} &    56.8\up{3.3} & 58.3\up{3.3} & 25.2\\

& & \textcolor[rgb]{0,0.5,0}{\Checkmark} & \textcolor[rgb]{0,0.5,0}{\Checkmark}& \textcolor[rgb]{1,0,0}{\XSolidBrush}& \textcolor[rgb]{1,0,0}{\XSolidBrush}& 59.2\up{5.7} & 63.7\up{7.1} & 41.1\up{15.7} \\

& & \textcolor[rgb]{0,0.5,0}{\Checkmark} & \textcolor[rgb]{1,0,0}{\XSolidBrush} &\textcolor[rgb]{0,0.5,0}{\Checkmark} &\textcolor[rgb]{1,0,0}{\XSolidBrush} & 57.3\up{3.8} & 60.5\up{3.9} & 24.6 \\

& & \textcolor[rgb]{0,0.5,0}{\Checkmark} & \textcolor[rgb]{0,0.5,0}{\Checkmark}& \textcolor[rgb]{0,0.5,0}{\Checkmark} & \textcolor[rgb]{1,0,0}{\XSolidBrush}& 62.1\up{8.6} & 65.8\up{9.2} & 45.2\up{19.8} \\

& & \textcolor[rgb]{0,0.5,0}{\Checkmark} & \textcolor[rgb]{1,0,0}{\XSolidBrush} & \textcolor[rgb]{0,0.5,0}{\Checkmark} &\textcolor[rgb]{0,0.5,0}{\Checkmark} & 58.6\up{5.1} & 61.3\up{4.7} & 25.7\up{0.5} \\

\rowcolor{ours} & & \textcolor[rgb]{0,0.5,0}{\Checkmark} & \textcolor[rgb]{0,0.5,0}{\Checkmark} & \textcolor[rgb]{0,0.5,0}{\Checkmark} & \textcolor[rgb]{0,0.5,0}{\Checkmark} & \textbf{65.4}\up{11.9} & \textbf{70.6}\up{14} & \textbf{52.3}\up{26.9}\\

\bottomrule
\end{tabular}
}
\vspace{-0.2cm}
\caption{
Ablation of dual-view and reasoning components.
}
\vspace{-0.25in}
\label{tab:4}
\end{table}

\noindent\textbf{Impact of GSXray Fine-tuning.}  
To further validate the effectiveness of the GSXray dataset, we fine-tuned representative open-source VLMs, including Qwen2.5-VL-7B \cite{Qwen2.5-VL}, InternVL3.5-8B \cite{wang2025internvl3_5}, and LLaVA-Next-7B \cite{liu2024llavanext}. All models exhibit substantial and consistent improvements across perception, relational, and occlusion tasks. Notably, LLaVA-OneVision-7B improves from 32.2 to 39.3 accuracy, and InternVL3.5-8B rises from 45.4 to 50.3, revealing remarkable performance gains. These results indicate that GSXray transfers domain knowledge effectively, enabling coherent cross-view geometric–semantic reasoning and substantially reducing the domain gap. 

% These results demonstrate that GSXray effectively transfers domain-specific knowledge, enabling general-purpose VLMs to reason geometrically and semantically across complementary X-ray views. The observed gains highlight GSXray’s pivotal role in bridging the domain gap and fostering robust dual-view understanding.

\subsection{Ablation Study}

\begin{table}[!t]
\Large
\vspace{-0.23in}
\centering
\resizebox{0.97\columnwidth}{!}{%
\begin{tabular}{l|cc|ccc}
\toprule
Model & GSXray & Dual view  &Acc & F1 & mIOU \\
\midrule
\multirow{2}{*}{Qwen2.5-VL-7B~\cite{Qwen2.5-VL}}&\multirow{2}{*}{\textcolor[rgb]{1,0,0}{\XSolidBrush}} &\textcolor[rgb]{1,0,0}{\XSolidBrush}  & 49.4 & 52.6  &  22.6  \\
& &\textcolor[rgb]{0,0.5,0}{\Checkmark} & 50.3$_{\color[rgb]{0.2,0.7,0.2}{\uparrow 0.9\%}}$ & 53.0 $_{\color[rgb]{0.2,0.7,0.2}{\uparrow 0.4\%}}$ &  20.3 $_{\color[rgb]{1,0.6,0.6}{\downarrow 2.3\%}}$ \\

\midrule
% \cline{2-2}
\multirow{2}{*}{InternVL3.5-8B~\cite{wang2025internvl3_5}} &\multirow{2}{*}{\textcolor[rgb]{1,0,0}{\XSolidBrush}}&\textcolor[rgb]{1,0,0}{\XSolidBrush}  & 43.2 & 50.3  &  18.6  \\
& &\textcolor[rgb]{0,0.5,0}{\Checkmark} & 45.4 $_{\color[rgb]{0.2,0.7,0.2}{\uparrow 2.2\%}}$& 51.5 $_{\color[rgb]{0.2,0.7,0.2}{\uparrow 1.0\%}}$ &  18.1 $_{\color[rgb]{1,0.6,0.6}{\downarrow 0.5\%}}$ \\
\midrule
% \cline{2-2}
\multirow{2}{*}{LLaVA-OneVision-7B~\cite{liu2023improvedllava}} &\multirow{2}{*}{\textcolor[rgb]{1,0,0}{\XSolidBrush}}&\textcolor[rgb]{1,0,0}{\XSolidBrush}  & 33.1 & 38.5  &  18.2  \\
& &\textcolor[rgb]{0,0.5,0}{\Checkmark} &32.3 $_{\color[rgb]{1,0.6,0.6}{\downarrow 0.8\%}}$& 36.8 $_{\color[rgb]{1,0.6,0.6}{\downarrow 1.7\%}}$ &  15.6 $_{\color[rgb]{1,0.6,0.6}{\downarrow 2.6\%}}$ \\

\midrule
% \midrule

\multirow{2}{*}{\textbf{InternVL3.5-8B (Ours)}} &\multirow{2}{*}{\textcolor[rgb]{0,0.5,0}{\Checkmark}}&\textcolor[rgb]{1,0,0}{\XSolidBrush}  & 45.2 & 49.6  &  20.5  \\
 & &\textcolor[rgb]{0,0.5,0}{\Checkmark} & 50.3$_{\color[rgb]{0.2,0.7,0.2}{\uparrow 5.1\%}}$ & 54.5 $_{\color[rgb]{0.2,0.7,0.2}{\uparrow 4.9\%}}$ &  23.2 $_{\color[rgb]{0.2,0.7,0.2}{\uparrow 2.7\%}}$\\
\midrule
\multirow{2}{*}{\textbf{LLaVA-OneVision-7B (Ours)}} &\multirow{2}{*}{\textcolor[rgb]{0,0.5,0}{\Checkmark}}&\textcolor[rgb]{1,0,0}{\XSolidBrush}  & 37.8 & 45.7  &  27.3  \\
 & &\textcolor[rgb]{0,0.5,0}{\Checkmark} & 39.3$_{\color[rgb]{0.2,0.7,0.2}{\uparrow 1.5\%}}$ & 47.2 $_{\color[rgb]{0.2,0.7,0.2}{\uparrow 1.5\%}}$ &  28.3 $_{\color[rgb]{0.2,0.7,0.2}{\uparrow 1.0\%}}$ \\
\midrule
 
\multirow{2}{*}{\textbf{GSR-8B (Ours)}} &\multirow{2}{*}{\textcolor[rgb]{0,0.5,0}{\Checkmark}}&\textcolor[rgb]{1,0,0}{\XSolidBrush}  & 62.1 & 67.4  & 49.6  \\
 && \textcolor[rgb]{0,0.5,0}{\Checkmark} & 65.4 $_{\color[rgb]{0.2,0.7,0.2}{\uparrow 3.3\%}}$& 70.6 $_{\color[rgb]{0.2,0.7,0.2}{\uparrow 3.2\%}}$ &  52.3 $_{\color[rgb]{0.2,0.7,0.2}{\uparrow 2.7\%}}$ \\

\bottomrule
\end{tabular}}
\vspace{-0.1in}
\caption{Ablation studies on the effect of the second view.}
\label{tab:abl2}
\vspace{-0.2in}
\end{table}

\noindent\textbf{Second View and Structured Reasoning.}
To evaluate the contributions of the second view and structured reasoning, we perform controlled ablations on DualXrayBench using Qwen3-VL-8B (Table~\ref{tab:4}). Using only the primary (top) view yields limited gains, highlighting that single-view perception is insufficient for spatial reasoning. Adding structured top-view descriptions (\texttt{<top>} with \texttt{<box>} grounding) improves performance, especially mIoU. Including the side view further enhances perceptual consistency and relational understanding, while dual views without textual guidance offer only marginal benefits and may reduce mIoU due to missing grounding. Full structured reasoning (\texttt{<top>}, \texttt{<side>}, \texttt{<conclusion>}, \texttt{<answer>}) with dual views achieves the strongest gains, forming GSR-8B. These results indicate that visual complementarity and structured reasoning are both essential, jointly enabling precise geometric alignment and robust cross-view inference.

\noindent\textbf{Role of the second view on DualXrayBench.} 
Table~\ref{tab:abl2} analyzes the influence of incorporating the second view on DualXrayBench before and after GSXray fine-tuning. Before training, the additional view yields inconsistent gains sometimes even degrading performance suggesting that vanilla VLMs fail to exploit cross-view cues, as unaligned side-view features act more as noise than complementary evidence. After fine-tuning, this behavior flips: removing the second view consistently harms Acc, F1, and mIoU. This shift verifies that GSXray effectively learns to extract and align semantic signals from the auxiliary view, converting an underutilized modality into a structured constraint that reinforces geometric grounding and cross-view reasoning.

\begin{table}[t]
    \centering
    \vspace{-0.2in}
    \resizebox{0.97\columnwidth}{!}{%
    \begin{tabular}{@{}lccccccccc@{}}
        \toprule
        Model & IL & CR & ID & IC & ML & IA & IT & Overall \\
        \midrule
        Florence-2 \cite{florenceV2} 
            & 30.1 & 37.5 & 39.8 & 30 & 21.1 & 35.8 & 29.1 & 32.3 \\
        MiniGPT \cite{minigptv2} 
            & 43.2 & 60.2 & 54.8 & 28.2 & 19.6 & 33.0 & 29.7 & 36.6 \\
        LLaVA 1.5 \cite{Llava15}  
            & 29.7 & 56.7 & 74.0 & 34.2 & 13.8 & 41.0 & 24.0 & 42.0 \\
        STING-BEE \cite{STING-BEE}
            & 49.2 & 79.2 & 80.0 & 45.2 & 27.8 & 53.0 & 35.0 & 52.8 \\
        \midrule
        \rowcolor{ours} \textbf{GSR-8B (Ours)} 
            & \textbf{52.5} & \textbf{83.0} & \textbf{83.5} & \textbf{47.0} & \textbf{29.5} & \textbf{57.0} & \textbf{41.1} & \textbf{55.3} \\
        \bottomrule
    \end{tabular}
}
    \vspace{-0.5em}
\caption{VQA performance comparison on \textbf{STING-BEE}.}
\label{tab:vqa_results}
% \vspace{-2em}
\end{table}
\subsection{Generalization}

\noindent\textbf{STING-BEE-VQA Evaluation.}
We further benchmark GSR-8B on the STING-BEE~\cite{STING-BEE} VQA benchmark, which evaluates multimodal reasoning across seven categories: \textit{Instance Location} (IL), \textit{Complex Reasoning} (CR), \textit{Instance Identity} (ID), \textit{Instance Counting} (IC), \textit{Misleading} (ML), \textit{Instance Attribute} (IA), and \textit{Instance Interaction} (IT). We fine-tune GSR-8B on 10k SFT data of STING-BEE and evaluate under the same multiple-choice format.
As reported in Table~\ref{tab:vqa_results}, GSR-8B consistently surpasses all vision-language baselines, showing the largest improvements in IL (+3.3), IC (+1.8), and CR (+3.8). These gains highlight stronger geometric–semantic reasoning and robustness to occlusion, showing that explicit geometric–semantic alignment enhances generalization across X-ray domains.

\noindent\textbf{Cross-Category and Domain Generalization.}
Following OVXD~\cite{ovxd}, we evaluate GSR-8B on PID under cross-category and cross-domain splits. \textit{Battery} (aligned with \textit{Power bank} in DualXrayCap) measures domain robustness, while the remaining unseen classes test category-level generalization. Under the standardized OVXD protocol for both VLMs and vision-only baselines, GSR-8B achieves the highest accuracy across all settings, confirming that structured cross-view reasoning and geometric–semantic alignment markedly improve transferability to novel categories and scanner shifts.
% Following the OVXD~\cite{ovxd} protocol, we evaluate the transferability of \textbf{GSR-8B} on the PID dataset under both cross-category and cross-domain settings. Four representative categories are used: \textit{Scissors}, \textit{Wrench}, \textit{Battery}, and \textit{Pliers}. Among them, \textit{Battery}—semantically aligned with \textit{Power bank} in DualXrayCap—tests cross-domain robustness, while the remaining unseen categories assess cross category generalization. All VLMs are evaluated strictly under the OVXD metrics and setup, and vision-only baselines are trained on base classes and tested on novel ones using the same protocol. As shown in Table~\ref{tab:5}, \textbf{GSR-8B} achieves the highest accuracy across both settings, indicating that structured cross-view reasoning and geometric–semantic alignment significantly enhance generalization to novel categories and scanner-induced domain shifts.

\begin{table}[!t]
\centering
\resizebox{0.97\columnwidth}{!}{%
\begin{tabular}{l|llllllllllllllllllllllll|c}
\toprule
Model & \multicolumn{6}{c}{Scissors} & \multicolumn{6}{c}{Wrench} & \multicolumn{6}{c}{Pliers} & \multicolumn{6}{c|}{Battery} & \textbf{AP50} \\
\midrule

OV-DETR \cite{ovdetr}& \multicolumn{6}{c}{0.2} & \multicolumn{6}{c}{0.1} & \multicolumn{6}{c}{0.1} & \multicolumn{6}{c|}{0.0} & 0.1 \\

RegionCLIP \cite{zhong2022regionclip}  & \multicolumn{6}{c}{0.4} & \multicolumn{6}{c}{2.6} & \multicolumn{6}{c}{2.4} & \multicolumn{6}{c|}{0.0} & 1.3 \\

BARON \cite{baron} & \multicolumn{6}{c}{1.5} & \multicolumn{6}{c}{18.1} & \multicolumn{6}{c}{3.1} & \multicolumn{6}{c|}{0.6} & 37.8 \\

\midrule
Faster RCNN \cite{ren2017faster}  & \multicolumn{6}{c}{0.0} & \multicolumn{6}{c}{0.2} & \multicolumn{6}{c}{0.1} & \multicolumn{6}{c|}{0.0} & 0.1 \\

YOLO11 \cite{yolo11_ultralytics} & \multicolumn{6}{c}{1.2} & \multicolumn{6}{c}{0.8} & \multicolumn{6}{c}{0.2} & \multicolumn{6}{c|}{0.1} & 2.3 \\
\midrule
OVXD \cite{ovxd}  & \multicolumn{6}{c}{16.9} & \multicolumn{6}{c}{46.2} & \multicolumn{6}{c}{6.4} & \multicolumn{6}{c|}{14.6} & 21.0 \\

\rowcolor{ours} \textbf{GSR-8B (Ours)} & \multicolumn{6}{c}{24.2} & \multicolumn{6}{c}{50.5} & \multicolumn{6}{c}{10.1} & \multicolumn{6}{c|}{20.4} & 26.3 \\
\bottomrule
\end{tabular}
}
\vspace{-0.15cm}
\caption{Cross-category and cross-domain generalization on \textbf{PID} following the OVXD protocol.}

\vspace{-0.45cm}
\label{tab:5}
\end{table}

\section{Conclusion} 
% In summary, DualXrayBench and GSXray establish a new paradigm for multi-view X-ray understanding. By training GSR to interpret the side view as a language-like constraint, we bridge geometric and semantic reasoning, achieving robust gains across tasks. We believe this framework offers a scalable path toward unified multimodal and multi-view learning.
In this work, we introduced DualXrayBench, the first comprehensive benchmark for X-ray inspection incorporating dual-view images and multimodal data, designed to evaluate cross-view reasoning. We presented a dual-view caption corpus and the GSXray dataset, which introduces structured Chain-of-Thought sequences to enable joint learning of cross-view geometry and cross-modal semantics. By leveraging these datasets, we proposed the GSR, which treats the second-view images as a ``language-like modality". Extensive evaluations on DualXrayBench show that GSR significantly improves performance across X-ray tasks, offering a new sight for real-world X-ray inspection applications. 

% This work paves the way for future advancements in multimodal X-ray detection, demonstrating the potential of dual-view images to enhance automated inspection systems.
% \input{sec/2_formatting}
% \input{sec/3_finalcopy}
{
    \small
    \bibliographystyle{ieeenat_fullname}
    \bibliography{main}

\begin{thebibliography}{48}
\providecommand{\natexlab}[1]{#1}
\providecommand{\url}[1]{\texttt{#1}}
\expandafter\ifx\csname urlstyle\endcsname\relax
  \providecommand{\doi}[1]{doi: #1}\else
  \providecommand{\doi}{doi: \begingroup \urlstyle{rm}\Url}\fi

\bibitem[Ahmed et~al.(2024)Ahmed, Velayudhan, Hassan, Bennamoun, Damiani, and Werghi]{BalAffin_Jour}
Abdelfatah Ahmed, Divya Velayudhan, Taimur Hassan, Mohammed Bennamoun, Ernesto Damiani, and Naoufel Werghi.
\newblock Enhancing security in x-ray baggage scans: A contour-driven learning approach for abnormality classification and instance segmentation.
\newblock \emph{Engineering Applications of Artificial Intelligence}, 130:\penalty0 107639, 2024.

\bibitem[Bai et~al.(2025)Bai, Chen, Liu, Wang, Ge, Song, Dang, Wang, Wang, Tang, Zhong, Zhu, Yang, Li, Wan, Wang, Ding, Fu, Xu, Ye, Zhang, Xie, Cheng, Zhang, Yang, Xu, and Lin]{Qwen2.5-VL}
Shuai Bai, Keqin Chen, Xuejing Liu, Jialin Wang, Wenbin Ge, Sibo Song, Kai Dang, Peng Wang, Shijie Wang, Jun Tang, Humen Zhong, Yuanzhi Zhu, Mingkun Yang, Zhaohai Li, Jianqiang Wan, Pengfei Wang, Wei Ding, Zheren Fu, Yiheng Xu, Jiabo Ye, Xi Zhang, Tianbao Xie, Zesen Cheng, Hang Zhang, Zhibo Yang, Haiyang Xu, and Junyang Lin.
\newblock Qwen2.5-vl technical report.
\newblock \emph{arXiv preprint arXiv:2502.13923}, 2025.

\bibitem[Bhowmik et~al.(2021)Bhowmik, Gaus, and Breckon]{deei6}
Neelanjan Bhowmik, Yona Falinie~A Gaus, and Toby~P Breckon.
\newblock On the impact of using x-ray energy response imagery for object detection via convolutional neural networks.
\newblock In \emph{2021 IEEE International Conference on Image Processing (ICIP)}, pages 1224--1228. IEEE, 2021.

\bibitem[Caldwell and Griffin(2019)]{compassxp}
Matthew Caldwell and Lewis~D Griffin.
\newblock Limits on transfer learning from photographic image data to x-ray threat detection.
\newblock \emph{Journal of X-ray Science and Technology}, 27\penalty0 (6):\penalty0 1007--1020, 2019.

\bibitem[Chen et~al.(2023)Chen, Zhu, Shen, Li, Liu, Zhang, Krishnamoorthi, Chandra, Xiong, and Elhoseiny]{minigptv2}
Jun Chen, Deyao Zhu, Xiaoqian Shen, Xiang Li, Zechun Liu, Pengchuan Zhang, Raghuraman Krishnamoorthi, Vikas Chandra, Yunyang Xiong, and Mohamed Elhoseiny.
\newblock Minigpt-v2: large language model as a unified interface for vision-language multi-task learning.
\newblock \emph{arXiv preprint arXiv:2310.09478}, 2023.

\bibitem[Chen et~al.(2024)Chen, Wu, Wang, Su, Chen, Xing, Zhong, Zhang, Zhu, Lu, et~al.]{chen2024internvl}
Zhe Chen, Jiannan Wu, Wenhai Wang, Weijie Su, Guo Chen, Sen Xing, Muyan Zhong, Qinglong Zhang, Xizhou Zhu, Lewei Lu, et~al.
\newblock Internvl: Scaling up vision foundation models and aligning for generic visual-linguistic tasks.
\newblock In \emph{Proceedings of the IEEE/CVF Conference on Computer Vision and Pattern Recognition}, pages 24185--24198, 2024.

\bibitem[DeepMind(2025{\natexlab{a}})]{gemini-2.5-flash}
Google DeepMind.
\newblock Gemini-2.5-flash.
\newblock \url{https://deepmind.google/models/gemini/flash/}, 2025{\natexlab{a}}.

\bibitem[DeepMind(2025{\natexlab{b}})]{gemini-2.5-pro}
Google DeepMind.
\newblock Gemini-2.5-pro.
\newblock \url{https://deepmind.google/models/gemini/pro/}, 2025{\natexlab{b}}.

\bibitem[He et~al.(2023)He, Mu, Ren, and Zhao]{LPIXray}
Chengquan He, Tong Mu, Weiping Ren, and Bohua Zhao.
\newblock Lpixray: A large-scale logistics prohibited item x-ray dataset for the application of deep learning in security inspection.
\newblock In \emph{2023 International Conference on Computers, Information Processing and Advanced Education (CIPAE)}, pages 481--485, 2023.

\bibitem[Hong et~al.(2025)Hong, Zhou, and Xu]{dagnet}
Shilong Hong, Yanzhou Zhou, and Weichao Xu.
\newblock Dagnet: A dual-view attention-guided network for efficient x-ray security inspection.
\newblock \emph{arXiv preprint arXiv:2502.01710}, 2025.

\bibitem[Isaac-Medina et~al.(2021)Isaac-Medina, Willcocks, and Breckon]{DBF}
Brian K.~S. Isaac-Medina, Chris~G. Willcocks, and Toby~P. Breckon.
\newblock Multi-view object detection using epipolar constraints within cluttered x-ray security imagery.
\newblock In \emph{2020 25th International Conference on Pattern Recognition (ICPR)}, pages 9889--9896, 2021.

\bibitem[Jocher and Qiu(2024)]{yolo11_ultralytics}
Glenn Jocher and Jing Qiu.
\newblock Ultralytics yolo11, 2024.

\bibitem[Kuckreja et~al.(2024)Kuckreja, Danish, Naseer, Das, Khan, and Khan]{geochat}
Kartik Kuckreja, Muhammad~Sohail Danish, Muzammal Naseer, Abhijit Das, Salman Khan, and Fahad~Shahbaz Khan.
\newblock Geochat: Grounded large vision-language model for remote sensing.
\newblock In \emph{Proceedings of the IEEE/CVF Conference on Computer Vision and Pattern Recognition}, pages 27831--27840, 2024.

\bibitem[Li et~al.(2024{\natexlab{a}})Li, Zhang, Guo, Zhang, Li, Zhang, Zhang, Zhang, Li, Liu, et~al.]{li2024llavaov}
Bo Li, Yuanhan Zhang, Dong Guo, Renrui Zhang, Feng Li, Hao Zhang, Kaichen Zhang, Peiyuan Zhang, Yanwei Li, Ziwei Liu, et~al.
\newblock Llava-onevision: Easy visual task transfer.
\newblock \emph{arXiv preprint arXiv:2408.03326}, 2024{\natexlab{a}}.

\bibitem[Li et~al.(2024{\natexlab{b}})Li, Wong, Zhang, Usuyama, Liu, Yang, Naumann, Poon, and Gao]{llava_med}
Chunyuan Li, Cliff Wong, Sheng Zhang, Naoto Usuyama, Haotian Liu, Jianwei Yang, Tristan Naumann, Hoifung Poon, and Jianfeng Gao.
\newblock Llava-med: Training a large language-and-vision assistant for biomedicine in one day.
\newblock \emph{Advances in Neural Information Processing Systems}, 36, 2024{\natexlab{b}}.

\bibitem[Li* et~al.(2022)Li*, Zhang*, Zhang*, Yang, Li, Zhong, Wang, Yuan, Zhang, Hwang, Chang, and Gao]{li2021glip}
Liunian~Harold Li*, Pengchuan Zhang*, Haotian Zhang*, Jianwei Yang, Chunyuan Li, Yiwu Zhong, Lijuan Wang, Lu Yuan, Lei Zhang, Jenq-Neng Hwang, Kai-Wei Chang, and Jianfeng Gao.
\newblock Grounded language-image pre-training.
\newblock In \emph{CVPR}, 2022.

\bibitem[Lin et~al.(2025{\natexlab{a}})Lin, Jia, Wang, Ma, and Li]{PIXray-Caption}
Shuyang Lin, Tong Jia, Hao Wang, Bowen Ma, and Mingyuan Li.
\newblock Open-vocabulary prohibited item detection for real-world x-ray security inspection.
\newblock \emph{IEEE Transactions on Information Forensics and Security}, 20:\penalty0 7469--7481, 2025{\natexlab{a}}.

\bibitem[Lin et~al.(2025{\natexlab{b}})Lin, Jia, Wang, Ma, Li, and Chen]{ovxd}
Shuyang Lin, Tong Jia, Hao Wang, Bowen Ma, Mingyuan Li, and Dongyue Chen.
\newblock Detection of novel prohibited item categories for real-world security inspection.
\newblock \emph{Engineering Applications of Artificial Intelligence}, 144:\penalty0 110110, 2025{\natexlab{b}}.

\bibitem[Liu et~al.(2023{\natexlab{a}})Liu, Li, Li, and Lee]{liu2023improvedllava}
Haotian Liu, Chunyuan Li, Yuheng Li, and Yong~Jae Lee.
\newblock Improved baselines with visual instruction tuning, 2023{\natexlab{a}}.

\bibitem[Liu et~al.(2023{\natexlab{b}})Liu, Li, Wu, and Lee]{liu2023llava}
Haotian Liu, Chunyuan Li, Qingyang Wu, and Yong~Jae Lee.
\newblock Visual instruction tuning, 2023{\natexlab{b}}.

\bibitem[Liu et~al.(2024{\natexlab{a}})Liu, Li, Li, and Lee]{Llava15}
Haotian Liu, Chunyuan Li, Yuheng Li, and Yong~Jae Lee.
\newblock Improved baselines with visual instruction tuning.
\newblock In \emph{Proceedings of the IEEE/CVF Conference on Computer Vision and Pattern Recognition}, pages 26296--26306, 2024{\natexlab{a}}.

\bibitem[Liu et~al.(2024{\natexlab{b}})Liu, Li, Li, Li, Zhang, Shen, and Lee]{liu2024llavanext}
Haotian Liu, Chunyuan Li, Yuheng Li, Bo Li, Yuanhan Zhang, Sheng Shen, and Yong~Jae Lee.
\newblock Llava-next: Improved reasoning, ocr, and world knowledge, 2024{\natexlab{b}}.

\bibitem[Ma et~al.(2022)Ma, Jia, Su, Jia, Chen, and Zhang]{PIXray}
Bowen Ma, Tong Jia, Min Su, Xiaodong Jia, Dongyue Chen, and Yichun Zhang.
\newblock Automated segmentation of prohibited items in x-ray baggage images using dense de-overlap attention snake.
\newblock \emph{IEEE Transactions on Multimedia}, 2022.

\bibitem[Ma et~al.(2024)Ma, Jia, Li, Wu, Wang, and Chen]{DvXray}
Bowen Ma, Tong Jia, Mingyuan Li, Songsheng Wu, Hao Wang, and Dongyue Chen.
\newblock Towards dual-view x-ray baggage inspection: A large-scale benchmark and adaptive hierarchical cross refinement for prohibited item discovery.
\newblock \emph{IEEE Transactions on Information Forensics and Security}, 2024.

\bibitem[Mery et~al.(2015)Mery, Riffo, Zscherpel, Mondrag{\'o}n, Lillo, Zuccar, Lobel, and Carrasco]{GDXray}
Domingo Mery, Vladimir Riffo, Uwe Zscherpel, German Mondrag{\'o}n, Iv{\'a}n Lillo, Irene Zuccar, Hans Lobel, and Miguel Carrasco.
\newblock Gdxray: The database of x-ray images for nondestructive testing.
\newblock \emph{Journal of Nondestructive Evaluation}, 34\penalty0 (4):\penalty0 42, 2015.

\bibitem[Miao et~al.(2019)Miao, Xie, Wan, Su, Liu, Jiao, and Ye]{SIXray}
Caijing Miao, Lingxi Xie, Fang Wan, Chi Su, Hongye Liu, Jianbin Jiao, and Qixiang Ye.
\newblock Sixray: A large-scale security inspection x-ray benchmark for prohibited item discovery in overlapping images.
\newblock In \emph{Proceedings of the IEEE/CVF conference on computer vision and pattern recognition}, pages 2119--2128, 2019.

\bibitem[OpenAI(2024)]{gpt4o}
OpenAI.
\newblock Openai-gpt-4o.
\newblock \url{https://openai.com/index/gpt-4o-system-card/}, 2024.

\bibitem[OpenAI(2025)]{o3}
OpenAI.
\newblock Openai-o3.
\newblock \url{https://openai.com/index/introducing-o3-and-o4-mini/}, 2025.

\bibitem[Radford et~al.(2021)Radford, Kim, Hallacy, Ramesh, Goh, Agarwal, Sastry, Askell, Mishkin, Clark, et~al.]{Clip_radford}
Alec Radford, Jong~Wook Kim, Chris Hallacy, Aditya Ramesh, Gabriel Goh, Sandhini Agarwal, Girish Sastry, Amanda Askell, Pamela Mishkin, Jack Clark, et~al.
\newblock Learning transferable visual models from natural language supervision.
\newblock In \emph{International conference on machine learning}, pages 8748--8763. PMLR, 2021.

\bibitem[Ren et~al.(2017)Ren, He, Girshick, and Sun]{ren2017faster}
Shaoqing Ren, Kaiming He, Ross~B Girshick, and Jian Sun.
\newblock Faster r-cnn: Towards real-time object detection with region proposal networks.
\newblock \emph{IEEE transactions on pattern analysis and machine intelligence}, 39\penalty0 (6):\penalty0 1137--1149, 2017.

\bibitem[Steitz et~al.(2018)Steitz, Saeedan, and Roth]{mvxray}
Jan-Martin~O Steitz, Faraz Saeedan, and Stefan Roth.
\newblock Multi-view x-ray r-cnn.
\newblock In \emph{German Conference on Pattern Recognition}, pages 153--168. Springer, 2018.

\bibitem[Tao et~al.(2021)Tao, Wei, Jiang, Li, Qin, Wang, Ma, Zhang, and Liu]{Hixray}
Renshuai Tao, Yanlu Wei, Xiangjian Jiang, Hainan Li, Haotong Qin, Jiakai Wang, Yuqing Ma, Libo Zhang, and Xianglong Liu.
\newblock Towards real-world x-ray security inspection: A high-quality benchmark and lateral inhibition module for prohibited items detection.
\newblock In \emph{Proceedings of the IEEE/CVF international conference on computer vision}, pages 10923--10932, 2021.

\bibitem[Tao et~al.(2022{\natexlab{a}})Tao, Li, Wang, Wei, Ding, Jin, Zhi, Liu, and Liu]{EDS}
Renshuai Tao, Hainan Li, Tianbo Wang, Yanlu Wei, Yifu Ding, Bowei Jin, Hongping Zhi, Xianglong Liu, and Aishan Liu.
\newblock Exploring endogenous shift for cross-domain detection: A large-scale benchmark and perturbation suppression network.
\newblock In \emph{2022 IEEE/CVF Conference on Computer Vision and Pattern Recognition (CVPR)}, pages 21157--21167. IEEE, 2022{\natexlab{a}}.

\bibitem[Tao et~al.(2022{\natexlab{b}})Tao, Wang, Wu, Liu, Liu, and Liu]{FSOD}
Renshuai Tao, Tianbo Wang, Ziyang Wu, Cong Liu, Aishan Liu, and Xianglong Liu.
\newblock Few-shot x-ray prohibited item detection: A benchmark and weak-feature enhancement network.
\newblock In \emph{Proceedings of the 30th ACM International Conference on Multimedia}, pages 2012--2020, 2022{\natexlab{b}}.

\bibitem[Tao et~al.(2025)Tao, Wang, Guo, Chen, Zhang, Liu, Wei, and Zhao]{ldxray}
Renshuai Tao, Haoyu Wang, Yuzhe Guo, Hairong Chen, Li Zhang, Xianglong Liu, Yunchao Wei, and Yao Zhao.
\newblock Dual-view x-ray detection: Can ai detect prohibited items from dual-view x-ray images like humans?
\newblock In \emph{Proceedings of the Computer Vision and Pattern Recognition Conference}, pages 10338--10347, 2025.

\bibitem[Team(2025)]{qwen3vl}
Qwen Team.
\newblock Qwen3 technical report, 2025.

\bibitem[Velayudhan et~al.(2025)Velayudhan, Ahmed, Alansari, Gour, Behouch, Hassan, Wasim, Maalej, Naseer, Gall, et~al.]{STING-BEE}
Divya Velayudhan, Abdelfatah Ahmed, Mohamad Alansari, Neha Gour, Abderaouf Behouch, Taimur Hassan, Syed~Talal Wasim, Nabil Maalej, Muzammal Naseer, Juergen Gall, et~al.
\newblock Sting-bee: Towards vision-language model for real-world x-ray baggage security inspection.
\newblock In \emph{Proceedings of the Computer Vision and Pattern Recognition Conference}, pages 20767--20777, 2025.

\bibitem[Wang et~al.(2021)Wang, Zhang, Wen, Liu, and Wu]{pidray_iccv}
Boying Wang, Libo Zhang, Longyin Wen, Xianglong Liu, and Yanjun Wu.
\newblock Towards real-world prohibited item detection: A large-scale x-ray benchmark.
\newblock In \emph{Proceedings of the IEEE/CVF international conference on computer vision}, pages 5412--5421, 2021.

\bibitem[Wang et~al.(2025)Wang, Gao, Gu, Pu, Cui, Wei, Liu, Jing, Ye, Shao, et~al.]{wang2025internvl3_5}
Weiyun Wang, Zhangwei Gao, Lixin Gu, Hengjun Pu, Long Cui, Xingguang Wei, Zhaoyang Liu, Linglin Jing, Shenglong Ye, Jie Shao, et~al.
\newblock Internvl3.5: Advancing open-source multimodal models in versatility, reasoning, and efficiency.
\newblock \emph{arXiv preprint arXiv:2508.18265}, 2025.

\bibitem[Wasim et~al.(2024)Wasim, Naseer, Khan, Yang, and Khan]{wasim2024vgdino}
Syed~Talal Wasim, Muzammal Naseer, Salman Khan, Ming-Hsuan Yang, and Fahad~Shahbaz Khan.
\newblock Videogrounding-dino: Towards open-vocabulary spatio-temporal video grounding.
\newblock In \emph{CVPR}, 2024.

\bibitem[Wei et~al.(2020)Wei, Tao, Wu, Ma, Zhang, and Liu]{OPIXray}
Yanlu Wei, Renshuai Tao, Zhangjie Wu, Yuqing Ma, Libo Zhang, and Xianglong Liu.
\newblock Occluded prohibited items detection: An x-ray security inspection benchmark and de-occlusion attention module.
\newblock In \emph{Proceedings of the 28th ACM international conference on multimedia}, pages 138--146, 2020.

\bibitem[Wu et~al.(2022)Wu, Yi, Zhang, Ouyang, and Yang]{dualray}
Modi Wu, Feifan Yi, Haigang Zhang, Xinyu Ouyang, and Jinfeng Yang.
\newblock Dualray: Dual-view x-ray security inspection benchmark and fusion detection framework.
\newblock In \emph{Chinese Conference on Pattern Recognition and Computer Vision (PRCV)}, pages 721--734. Springer, 2022.

\bibitem[Wu et~al.(2023)Wu, Zhang, Jin, Liu, and Loy]{baron}
Size Wu, Wenwei Zhang, Sheng Jin, Wentao Liu, and Chen~Change Loy.
\newblock Aligning bag of regions for open-vocabulary object detection.
\newblock In \emph{Proceedings of the IEEE/CVF conference on computer vision and pattern recognition}, pages 15254--15264, 2023.

\bibitem[Xiao et~al.(2024)Xiao, Wu, Xu, Dai, Hu, Lu, Zeng, Liu, and Yuan]{florenceV2}
Bin Xiao, Haiping Wu, Weijian Xu, Xiyang Dai, Houdong Hu, Yumao Lu, Michael Zeng, Ce Liu, and Lu Yuan.
\newblock Florence-2: Advancing a unified representation for a variety of vision tasks.
\newblock In \emph{Proceedings of the IEEE/CVF Conference on Computer Vision and Pattern Recognition}, pages 4818--4829, 2024.

\bibitem[Zang et~al.(2022)Zang, Li, Zhou, Huang, and Loy]{ovdetr}
Yuhang Zang, Wei Li, Kaiyang Zhou, Chen Huang, and Chen~Change Loy.
\newblock Open-vocabulary detr with conditional matching.
\newblock In \emph{European conference on computer vision}, pages 106--122. Springer, 2022.

\bibitem[Zhao et~al.(2022)Zhao, Zhu, Dou, Deng, and Wang]{CLCXray}
Cairong Zhao, Liang Zhu, Shuguang Dou, Weihong Deng, and Liang Wang.
\newblock Detecting overlapped objects in x-ray security imagery by a label-aware mechanism.
\newblock \emph{IEEE transactions on information forensics and security}, 17:\penalty0 998--1009, 2022.

\bibitem[Zhong et~al.(2022)Zhong, Yang, Zhang, Li, Codella, Li, Zhou, Dai, Yuan, Li, et~al.]{zhong2022regionclip}
Yiwu Zhong, Jianwei Yang, Pengchuan Zhang, Chunyuan Li, Noel Codella, Liunian~Harold Li, Luowei Zhou, Xiyang Dai, Lu Yuan, Yin Li, et~al.
\newblock Regionclip: Region-based language-image pretraining.
\newblock In \emph{Proceedings of the IEEE/CVF conference on computer vision and pattern recognition}, pages 16793--16803, 2022.

\bibitem[Zhu et~al.(2025)Zhu, Wang, Chen, Liu, Ye, Gu, Tian, Duan, Su, Shao, et~al.]{zhu2025internvl3}
Jinguo Zhu, Weiyun Wang, Zhe Chen, Zhaoyang Liu, Shenglong Ye, Lixin Gu, Hao Tian, Yuchen Duan, Weijie Su, Jie Shao, et~al.
\newblock Internvl3: Exploring advanced training and test-time recipes for open-source multimodal models.
\newblock \emph{arXiv preprint arXiv:2504.10479}, 2025.

\end{thebibliography}
}


\end{document}